\documentclass[11pt,a4paper]{article}
\usepackage[margin=25mm,top=25mm,headheight=41pt,headsep=8pt]{geometry}
\usepackage{lmodern}

\usepackage{graphicx}
\usepackage[hyphens]{url}
\usepackage{amsmath,amssymb,mathtools}
\usepackage{bm}
\newcommand{\mathds}{\mathbb}
\usepackage{booktabs}
\usepackage{array}
\usepackage{pdflscape}
\usepackage{algorithm,algpseudocode}
\usepackage{setspace}
\usepackage{enumitem}
\usepackage{subcaption}
\usepackage{xcolor}
\usepackage{tcolorbox}
\tcbuselibrary{breakable}
\usepackage{fontawesome5}

\usepackage[american]{babel}
\usepackage{csquotes}
\usepackage[style=apa,backend=biber,natbib=true]{biblatex}
\usepackage[hidelinks]{hyperref}
\newcommand{\paperref}[2]{\hyperref[#2]{#1~\ref*{#2}}}

\usepackage[section]{placeins}
\pretocmd{\subsection}{\FloatBarrier}{}{}
\usepackage{styles/lab-preprint}
\newcommand{\PaperTitle}{Dynamic Soft Embodiment Grouping for Cross-Embodiment Offline Reinforcement Learning}
\newcommand{\PaperAuthors}{AmirAli Fakhari Zavareh\textsuperscript{a}, Jianming Ma\textsuperscript{a},
Zhenzhe Sun\textsuperscript{b}\thanks{ORCID: \href{https://orcid.org/0009-0003-8799-8079}{0009-0003-8799-8079}.},
Yue Gao\textsuperscript{a}\thanks{Corresponding author.}}
\newcommand{\PaperAffiliation}{\textsuperscript{a}Shanghai Jiao Tong University\\
\small No. 800 Dongchuan Road, Minhang District, Shanghai, 200240, China\\
\small\textsuperscript{b}ZTE Corporation}
\newcommand{\PaperContact}{amiralifakhari@sjtu.edu.cn, nymjm11@sjtu.edu.cn\\
\small\ttfamily sun.zhenzhe5@zte.com.cn, yuegao@sjtu.edu.cn}
\renewcommand{\LabName}{RL\textsuperscript{2} Lab}
\renewcommand{\LabLogoPath}{assets/lab-logo.png}

\newcommand{\sep}{; }
\hypersetup{pdftitle={\PaperTitle},pdfauthor={AmirAli Fakhari Zavareh, Jianming Ma, Zhenzhe Sun, Yue Gao}}
\makeatletter
\let\originalappendix\appendix
\renewcommand{\appendix}{%
  \originalappendix
  \def\p@section{Appendix~}%
  \def\p@subsection{Appendix~}%
  \counterwithin{equation}{section}%
  \renewcommand{\thetable}{\thesection.\arabic{table}}%
  \renewcommand{\thefigure}{\thesection.\arabic{figure}}%
}

\makeatother

\newtcolorbox{degkeybox}[1]{
    breakable,
    colback=gray!6,
    colframe=gray!55,
    colbacktitle=gray!16,
    coltitle=black,
    fonttitle=\bfseries\small,
    boxrule=0.4pt,
    arc=1.5mm,
    left=6pt, right=6pt, top=4pt, bottom=4pt,
}

\begin{document}
\title{\bfseries\PaperTitle}
\author{\PaperAuthors\\[4pt]\small\PaperAffiliation\\[4pt]\small\ttfamily\PaperContact}
\date{}
\maketitle
\thispagestyle{labfirst}
\begin{abstract}
Cross-embodiment offline reinforcement learning offers a potential way to address the limited availability of robot interaction data. It aims to learn a shared policy from datasets collected across robots with different physical configurations. However, these differences can lead to conflicting updates during shared actor training. Existing Embodiment Grouping (EG) uses fixed groups based on robot morphology and does not update their assignments as actor gradients change. We introduce Dynamic Embodiment Grouping (DEG), which periodically combines morphology and current actor-gradient information to refresh groups and compute soft robot contributions to sequential group updates. Across 16 simulated robots, we compare 31 experimental configurations that vary grouping, gradient processing, and policy architecture under a common evaluation protocol. In the evaluated grouping comparisons, soft assignments yield higher mean returns than hard assignments when morphology and gradients are combined; this ordering reverses for gradient-only grouping. Across five seeds on the evaluated 70\% Suboptimal Forward dataset, DEG achieves a mean return of 52.79, 4.51 points higher than static EG (48.28), and a standard deviation of 2.13, compared with 5.88 for static EG. However, this improvement is not consistently observed on expert and replay datasets, suggesting that DEG's effectiveness depends on the data regime being evaluated.

\end{abstract}
\noindent\textbf{Keywords: }Cross-embodiment learning \sep Offline reinforcement learning \sep Embodiment grouping \sep Gradient conflict \sep Multi-task learning \sep Robot locomotion

\section{Introduction}

Robot experience is expensive to collect: trajectories require physical platforms, safe interaction, task-specific instrumentation, and often expert teleoperation, which can make training a separate effective policy for every embodiment impractical. Cross-embodiment learning addresses this by pooling data across platforms, as demonstrated by Open X-Embodiment and Octo~\citep{open_x_embodiment2024,octo2024}. Related work on generalist policies, such as RT-2, combines robot trajectories with vision-language pretraining to improve robotic control~\citep{brohan2023rt2}. Offline reinforcement learning (RL) further learns from fixed datasets that include suboptimal as well as expert data~\citep{kostrikov2022iql,springenberg2024offline}.

Pooling heterogeneous data, however, does not guarantee that every robot benefits from shared training. Robots differ in their dynamics, action spaces, and data distributions; an actor update that improves one embodiment may hinder another, producing negative transfer~\citep{yu2020pcgrad}. In IQL, the actor is trained through advantage-weighted behavioral cloning, so each dataset action's estimated advantage determines how strongly it contributes to actor updates~\citep{kostrikov2022iql}. Although the actor parameters are shared, each embodiment's contribution to the update is shaped by its dataset and learned advantage estimates. This raises the question of how robot-specific learning signals should be organized when updating a shared policy. We examine this question with particular attention to learning from datasets that include suboptimal behavior.

Embodiment Grouping (EG) addresses cross-robot interference by representing robots as morphology graphs, assigning them to fixed hard groups based on structural similarity, and applying sequential group-wise actor updates~\citep{abe2024eg}. \citet[\href{https://proceedings.iclr.cc/paper_files/paper/2026/file/24a0f56452202e8433d582dbf5095c32-Paper-Conference.pdf\#page=7}{Section~5.2}]{abe2024eg} report a positive correlation between morphological similarity and policy-gradient alignment (Pearson $r=0.63$), indicating that morphology is informative about gradient relationships in their evaluated setting. However, this correlation does not establish that morphology explains all variation in pairwise gradient alignment, or that these relationships remain unchanged as the policy parameters evolve. In broader multi-task learning, training-derived task affinities have been used to select task groupings and to update task groups dynamically during training~\citep{fifty2021tag,jeong2025stgu}. These methods establish relevant alternatives, but leave open a distinct empirical question in the cross-embodiment offline-RL setting: how should structural information and current actor-gradient relationships jointly determine robot contributions to group updates?

We use the URMA shared-actor architecture~\citep{bohlinger2024urma} and IQL objectives~\citep{kostrikov2022iql}, with static EG as the main baseline~\citep{abe2024eg}. We introduce Dynamic Embodiment Grouping (DEG), which periodically combines morphology and current actor-gradient information to compute soft contributions from robots to sequential group updates. On the primary 70\% Suboptimal Forward dataset, DEG achieves a higher mean return than static EG (52.79 vs. 48.28 across five seeds), with lower seed-to-seed variability (standard deviations 2.13 vs. 5.88). This advantage does not hold uniformly across the expert and replay datasets. We also evaluate two DEG ablations that separately remove continued regrouping and group-specific effects (\paperref{Section}{sec:deg_ablations}), alongside alternative grouping, gradient-processing, and policy-architecture configurations. \paperref{Section}{sec:setup} describes the common evaluation protocol and its difference from the protocol reported in the EG work~\citep{abe2024eg}.

\begin{degkeybox}{\faLightbulb[regular]\hspace{0.6em}Key Contributions}
    Our contributions are:
    \begin{itemize}[leftmargin=1.2em, itemsep=2pt, topsep=3pt]
        \item We describe DEG, a morphology- and gradient-informed soft-contribution scheme for sequential group-wise actor updates in cross-embodiment offline RL.
        \item We compare update configurations under a common protocol on a 16-robot simulated locomotion benchmark, covering grouping, gradient-processing, and architecture variants, and report aggregate and per-robot outcomes with variability across random seeds.
        \item We compare DEG with static EG across suboptimal, expert, and replay data conditions, documenting how relative performance differs across data conditions, including settings in which DEG does not improve static EG.
    \end{itemize}
    \tcblower
    {\small\textbf{Project page}\enspace\url{https://rl2-theory.github.io/deg-project-page/}}
\end{degkeybox}

\section{Related Work}

\subsection{Cross-Embodiment Offline RL and Embodiment Grouping}

Cross-embodiment learning has developed through shared robot datasets and policies trained across platforms. Open X-Embodiment provides standardized heterogeneous robot datasets and demonstrates transfer across robots with its RT-X models~\citep{open_x_embodiment2024}. Octo builds on these datasets to pretrain a generalist manipulation policy that can be fine-tuned to new observation and action spaces~\citep{octo2024}. These efforts provide data and policy infrastructure for reusing experience across embodiments.

Offline RL offers a complementary approach to learning from recorded experience by using rewards to learn policies from fixed datasets, including suboptimal behavior. IQL learns value functions from dataset actions and extracts a policy through advantage-weighted behavioral cloning~\citep{kostrikov2022iql}. IQL supplies an actor objective but does not introduce an embodiment-grouping mechanism for organizing shared-policy updates across robots.

Abe et al.~\citep{abe2024eg} systematically investigate cross-embodiment offline RL on a 16-robot locomotion benchmark. Their experiments show that offline RL can outperform behavior cloning on datasets rich in suboptimal trajectories, but also identify gradient interference as the proportion of suboptimal data and the number of robot types increase. Embodiment Grouping (EG) is the direct foundation for our setting: it applies cross-embodiment offline RL to heterogeneous robot datasets, groups robots using distances between morphology graphs, and performs sequential group-wise actor updates with a fixed hard partition~\citep{abe2024eg}.

A fixed partition does not adapt group assignments to current actor-gradient relationships. The question addressed here is whether morphology and current learning signals can jointly guide robot contributions as training proceeds. DEG extends this framework with periodic regrouping and soft robot contributions informed by current actor gradients as well as morphology.

\subsection{Task Relationships, Grouping, and Gradient Conflict}

Several multi-task methods study how training relationships can inform task grouping. Task Affinity Grouping (TAG) estimates how one task's gradient update changes other tasks' losses, then uses the resulting affinities to select task groupings for separate multi-task networks~\citep{fifty2021tag}. Grad-TAG develops a gradient-based estimator of task affinity for scalable grouping~\citep{li2024gradtag}. These methods establish that training-derived relationships can inform grouping, but TAG's goal is selecting networks to serve tasks, rather than assigning robot contributions to sequential updates of one shared policy.

Selective Task Group Updates (STGU) is more closely related to DEG's update procedure~\citep{jeong2025stgu}. STGU tracks proximal inter-task affinity during optimization, adaptively forms disjoint groups, and updates these groups sequentially within training. STGU uses proximal task affinity to form discrete task groups, whereas DEG combines morphology and actor-gradient information to compute robot-to-group contributions, including soft assignments. DEG's scope is this formulation in cross-embodiment offline RL.

Gradient-conflict methods provide a complementary approach to organizing updates. PCGrad projects conflicting task gradients before aggregation~\citep{yu2020pcgrad}, while CAGrad chooses an aggregate update that balances average improvement with worst-task local improvement~\citep{liu2021cagrad}. Unlike grouping methods, these methods modify gradient aggregation without constructing robot groups.

Other methods address task-update weighting and optimizer dynamics. Pascal et al. analyze alternating task-specific updates, show that moving-average optimizers can carry update history across task-specific steps, and propose separate optimizer states for each task. They also introduce random task grouping to trade off task separation and computational cost~\citep{pascal2021improved}. DB-MTL addresses task balancing by transforming losses and normalizing gradient magnitudes~\citep{lin2026dbmtl}. These methods are relevant to DEG's sequential updates, but address different mechanisms: separate optimizer states control how update history is retained, while DB-MTL balances loss scales and gradient magnitudes. DEG instead computes morphology- and gradient-informed robot-to-group contributions.

\subsection{Architectures for Heterogeneous Robots}

Grouping determines how training signals are combined, while architecture determines how parameters and representations are shared. URMA uses descriptor-conditioned attention to aggregate variable-length joint and foot observation sets into fixed-dimensional representations~\citep{bohlinger2024urma}. Our shared policy uses this encoder design. MOORE uses a mixture-of-orthogonal-experts design for multi-task reinforcement learning~\citep{hendawy2024moore}. We include architecture variants as exploratory configuration comparisons. In particular, the MOORE-labelled implementation and its result should be interpreted as the implementation evaluated here, not as a comprehensive evaluation of the original method.

Beyond the update framework studied here, recent cross-embodiment policy work develops complementary directions: CrossFormer trains a single transformer policy on 900K trajectories collected from 20 robot embodiments~\citep{doshi2025scaling}, and Multi-Loco combines a morphology-agnostic diffusion model with a lightweight residual policy optimized by reinforcement learning across legged embodiments~\citep{yang2025multiloco}. These directions modify policy architectures or generative priors; DEG instead organizes the updates of a shared policy within cross-embodiment offline RL.

\section{Problem and Backbone}\label{sec:backbone}

We first define the learning setting and the shared actor loss, then describe the grouped updates inherited from EG. These establish the training framework within which DEG changes robot contributions to actor updates.

\subsection{Learning Setting and Data}
Let $\mathcal{T}=\{1,\dots,N\}$ denote the set of $N$ robot embodiments. Each embodiment $\tau$ has its own Markov decision process $\mathcal{M}_{\tau}$ and a fixed offline dataset $\mathcal{D}_{\tau}$. We train a single policy $\pi_{\theta}(a|s,\tau)$, conditioned on embodiment descriptors that encode each robot's physical and actuation properties, such as joint limits, actuator limits, and link geometry, to maximize average return:
\begin{equation}
\max_{\theta} J(\theta)=\sum_{\tau\in\mathcal{T}}p(\tau)\,\mathbb{E}_{\pi_{\theta},\mathcal{M}_{\tau}}\left[\sum_{t=0}^{\infty}\gamma^t r_{\tau}(s_t,a_t)\right],
\label{eq:ce_objective}
\end{equation}
Here $\theta$ denotes the shared actor parameters, $s_t$ and $a_t$ are the state and action at environment timestep $t$, and $r_{\tau}(s_t,a_t)$ is the reward for embodiment $\tau$. The objective $J(\theta)$ averages discounted return across embodiments: $\gamma\in(0,1)$ is the discount factor, and $p(\tau)=1/N$ in our experiments. Training uses only the fixed datasets. The formal MDP and transition-record definitions are provided in~\ref{app:methods}.

The EG locomotion dataset suite~\citep{abe2024eg} includes Forward and Backward variants across 16 simulated robots: nine quadrupeds, six bipeds, and one hexapod. Our experiments use its three Forward variants. The evaluated task is to track a forward velocity of $1$\,m/s on flat terrain in MuJoCo, using the locomotion reward from URMA~\citep{todorov2012mujoco,bohlinger2024urma}. The evaluated datasets are:
\begin{itemize}[leftmargin=*,nosep]
    \item \textbf{Expert Forward:} expert PPO data.
    \item \textbf{Replay Forward:} uniformly sampled PPO training data covering the course of training.
    \item \textbf{70\% Suboptimal Forward:} 70\% early suboptimal PPO data and 30\% later expert-like PPO data.
\end{itemize}
The locally available Suboptimal and Replay variants each contain 1M transitions per robot. We use 70\% Suboptimal Forward as the primary evaluation set and the Expert and Replay variants for cross-dataset analysis. Collection and split construction follow~\citet{abe2024eg}; the reward components and naming correspondence with the EG paper are documented in~\ref{app:methods}. The released datasets are available at:\par
{\footnotesize\url{https://huggingface.co/datasets/haruki-abe/cross_embodiment_offline_rl}}\par

\subsection{Shared Policy and Actor Objective}
The main configuration uses URMA observation encoders to map general observations and variable-length joint and foot inputs into a shared representation~\citep{bohlinger2024urma}. The actor, value function, and each Q-function have separate parameter sets, with each set shared across robots. Embodiment descriptors and robot-specific observations condition their outputs. Shared actor output projections operate on joint-specific features to produce the action distributions for each robot's valid action coordinates. Thus robots with different observation and action dimensions can train the same actor. The encoder decomposition is given in~\ref{app:a_backbone}, and network widths and training settings are listed in~\ref{app:training}.

We train this backbone with Implicit Q-Learning (IQL)~\citep{kostrikov2022iql}. The value and Q-functions learn from dataset transitions, while the actor uses advantage-weighted behavioral cloning. The actor loss is central to our method because its gradient supplies each robot's contribution to the shared update. For embodiment $\tau$, it is
\begin{align}
\mathcal{L}^{\pi}_{\tau}(\theta) &= -\mathbb{E}_{(s,a)\sim\mathcal{D}_{\tau}}\left[w(s,a,\tau)\,\ell_{\theta}(a|s,\tau)\right], \label{eq:iql_pi}\\
\ell_{\theta}(a|s,\tau) &= \frac{1}{n_{\tau}}\sum_{k=1}^{n_{\tau}}\log\pi_{\theta,k}(a_k|s,\tau),\\
w(s,a,\tau) &= \min\left(\exp\left(\beta A(s,a,\tau)\right),c_A\right),\\
A(s,a,\tau) &= \min(Q_{\phi_1},Q_{\phi_2})(s,a,\tau)-V_{\psi}(s,\tau).\nonumber
\end{align}
The policy is a factorized Gaussian. Its log-likelihood $\ell_{\theta}$ averages over the $n_{\tau}$ valid action coordinates; $\pi_{\theta,k}$ is the density for coordinate $k$, and padded coordinates are excluded. The advantage $A$ is the minimum of the two online critic estimates, $Q_{\phi_1}$ and $Q_{\phi_2}$, minus the value baseline $V_{\psi}$; $\phi_1,\phi_2$ and $\psi$ denote their respective parameter sets. The positive coefficient $\beta$ controls how strongly advantage differences affect the action weight $w(s,a,\tau)$, and $c_A$ caps that weight. Their experimental values are listed in \paperref{Table}{tab:hyperparameters}. The expectile-value and twin-Q losses are given in~\ref{app:a_backbone}; the target-network updates are specified in \paperref{Algorithm}{alg:eg}.

\subsection{Grouped Actor Updates in EG}\label{sec:eg_training}
Static EG is the main baseline~\citep{abe2024eg}. It represents robots as morphology graphs, computes fused Gromov--Wasserstein (FGW) dissimilarities between these graphs~\citep{vayer2019fgw}, and clusters the robots into a fixed partition before training. At each training iteration, one minibatch is sampled per robot. The value and Q losses are averaged across these minibatches, while the actor receives a sequence of group updates. Each group update uses the unweighted mean of the actor gradients of its members.

These actor updates share one Adam optimizer~\citep{kingma2015adam}, whose state carries between group steps. Groups are visited in the same pseudorandom order at every iteration. This grouped actor-update framework is the starting point for DEG: the shared backbone and global critic losses are retained, while group formation and robot contributions change. The full static-EG equations, graph construction, and clustering qualifications are provided in~\ref{app:methods}.

\section{Dynamic Soft Embodiment Grouping}

DEG changes two parts of static EG: it refreshes the groups during training and replaces exclusive group membership with soft contributions. A refresh first measures relationships between current actor gradients, combines this information with morphology, and clusters the resulting robot distances. The clusters then define the neighborhoods used to calculate robot-to-group weights. Each group update combines all robot gradients according to these weights.

There are two timescales. Actor gradients are computed at every training iteration, while the groups and contribution weights are refreshed periodically and held fixed between refreshes. We describe the main FGW configuration below, followed by the comparison configurations and ablations.

\subsection{Dynamic Embodiment Grouping (DEG)}\label{sec:exp_grouping}

\subsubsection{Distances and Group Formation}
For the remaining description, $i$ and $j$ index embodiments in $\mathcal{T}$; they play the same role as $\tau$ in \paperref{Section}{sec:backbone}. At a refresh, we compute the actor gradient $\bm{g}_i=\nabla_{\theta}\mathcal{L}^{\pi}_i(\theta)$ from one minibatch per robot, evaluating all gradients at the same current actor parameters. Their cosine similarity is
\begin{equation}
S_{ij}=\frac{\bm{g}_i^{\top}\bm{g}_j}{\|\bm{g}_i\|_2\|\bm{g}_j\|_2+\epsilon_{\mathrm{cos}}},
\label{eq:cosine_sim}
\end{equation}
where the Euclidean norms are computed on flattened parameter gradients, the positive constant $\epsilon_{\mathrm{cos}}$ prevents division by zero, and the diagonal is explicitly set to $S_{ii}=1$.

Morphology enters through a kinematic graph $\mathcal{G}_i$ for each robot. Nodes carry features from the robot description, and the structural matrix contains geometry-weighted shortest-path distances~\citep{dijkstra1959note}. We augment every node with two robot-level statistics so that the graph comparison also reflects current learning signals:
\begin{equation}
\mathbf{x}'_{iv}=\left[\mathbf{x}_{iv},\;b_1\bar{S}_i,\;b_2\Delta\bar{S}_i\right], \qquad
\bar{S}_i=\frac{1}{N}\sum_{j=1}^{N}S_{ij},
\label{eq:aug_features}
\end{equation}
Here $\mathbf{x}_{iv}$ is the original feature vector of node $v$ in robot $i$, and $\bar{S}_i$ averages similarity over all robots, including $i$ itself. The change $\Delta\bar{S}_i$ is measured since the preceding refresh and is zero at the first refresh. The coefficients $b_1$ and $b_2$ scale the appended mean and change statistics, respectively.

FGW dissimilarity compares the node features and structural matrices of two graphs~\citep{vayer2019fgw}. We evaluate $d_{\mathrm{FGW}}(\mathcal{G}_i,\mathcal{G}_j)$ using the augmented features above. Consequently, this graph dissimilarity contains both morphology and summary gradient information. We also include the pairwise gradient dissimilarity $1-S_{ij}$ directly. The two terms are normalized separately and blended:
\begin{equation}
\begin{aligned}
D_{ij} ={} & (1-\alpha_g)\,\frac{d_{\mathrm{FGW}}(\mathcal{G}_i,\mathcal{G}_j)}{\max_{u,v} d_{\mathrm{FGW}}(\mathcal{G}_u,\mathcal{G}_v)+\epsilon_{\mathrm{norm}}} \\
&+ \alpha_g\,\frac{1 - S_{ij}}{\max_{u,v}(1 - S_{uv})+\epsilon_{\mathrm{norm}}},
\end{aligned}
\label{eq:blend}
\end{equation}
The maxima range over all robot pairs $u,v$. The coefficient $\alpha_g\in[0,1]$ controls the balance between graph and pairwise gradient dissimilarities, and the positive constant $\epsilon_{\mathrm{norm}}$ stabilizes their normalization. Ward hierarchical clustering~\citep{ward1963hierarchical} of $D$ produces $M$ groups $\{G_m\}_{m=1}^{M}$, where $G_m$ is the set of robots in group $m$. The full FGW objective, graph construction, and Ward-linkage qualification are given in~\ref{app:a_static_eg} and~\ref{app:a_deg}.

\subsubsection{From Groups to Actor Updates}\label{sec:soft_weights}
The groups determine which robot neighborhoods define each update. They do not restrict contributions to robots inside the group: any robot can contribute according to its affinity to the group members. For robot $i$ and group $G_m$, we sum the pairwise affinities and normalize across groups:
\begin{align}
A_{im} &= \sum_{j\in G_m}\exp(-D_{ij}/\kappa), \\
w_{im} &= \frac{A_{im}}{\sum_{m'=1}^{M}A_{im'}+\epsilon_w},
\label{eq:soft_weights}
\end{align}
Here $A_{im}$ is the group affinity, and the positive temperature $\kappa$ controls its sensitivity to distance. The positive constant $\epsilon_w$ stabilizes the normalization denominator. Smaller distances give larger affinities. The weight $w_{im}$ therefore expresses robot $i$'s contribution to group $m$ relative to its contributions to the other groups. These robot-to-group weights differ from the dataset-action weights $w(s,a,\tau)$ in the IQL loss; likewise, $A_{im}$ denotes group affinity rather than the action advantage $A(s,a,\tau)$.

The resulting matrix $\mathbf{W}=[w_{im}]\in\mathbb{R}^{N\times M}$ has one row per robot and one column per group. It is computed at each successful refresh and held fixed until the next one. During ordinary training iterations, group $m$ uses these weights to form
\begin{equation}
\bm{g}^{(m)} = \frac{\sum_{i=1}^{N} w_{im}\, \bm{g}_i}{\sum_{i=1}^{N} w_{im} + \epsilon_g},
\label{eq:soft_grad}
\end{equation}
where the positive constant $\epsilon_g$ stabilizes the denominator. This denominator normalizes the total robot mass assigned to that update. Each $\bm{g}^{(m)}$ is passed sequentially to the shared Adam optimizer. All per-robot gradients are computed once at the same actor parameters and reused across the $M$ group steps; they are not recomputed after each step. Thus a robot may contribute to several sequential updates, with different weights in each. With weights fixed, this aggregate is the gradient of a normalized weighted actor loss; its explicit expression is given in~\ref{app:methods}.

\subsubsection{Initialization and Refresh Schedule}
Training starts from the static-EG morphology partition with uniform weights $w_{im}=1/M$. These weights produce identical group gradients until the first successful refresh. Refreshes are attempted at multiples of the interval $K$ once the warmup of $W_0$ iterations is complete. At a refresh, robot gradients are recomputed at the post-update actor parameters, and the distances, groups, and weights are recalculated. The refresh is skipped if the largest robot-gradient norm is below the threshold $\delta_g$; otherwise, the new groups and weights take effect in the next iteration. Between refreshes, $\mathbf{W}$ remains fixed while the actor gradients continue to change.

The value and twin Q-functions retain the IQL losses averaged across all robots. \paperref{Algorithm}{alg:deg_overview} summarizes the update order. Gradient clipping, learning-rate scheduling, numerical safeguards, and the full implementation algorithm are documented in~\ref{app:a_deg}; the numerical settings are collected in \paperref{Table}{tab:hyperparameters}.

\begin{algorithm}[H]
\caption{DEG overview (FGW configuration)}
\label{alg:deg_overview}
\begin{algorithmic}[1]
\small\singlespacing
\Require Robot datasets and group count $M$
\Require Warmup $W_0$, refresh interval $K$, gradient threshold $\delta_g$
\State Initialize static-EG groups; set $w_{im}\gets1/M$
\For{each training iteration $t$}
    \State Sample one minibatch per robot; update the IQL value function
    \State Compute $\bm{g}_i\gets\nabla_{\theta}\mathcal{L}^{\pi}_i(\theta)$ for all robots at the same actor parameters
    \For{each group $m$ in the fixed update order}
        \State Form $\bm{g}^{(m)}$ using \paperref{Eq.}{eq:soft_grad}; apply a shared-Adam actor step
    \EndFor
    \State Update the twin Q-functions and their targets
    \If{$t\geq W_0$ and $t\bmod K=0$}
        \State Recompute robot gradients at the post-update actor parameters
        \If{the largest gradient norm is at least $\delta_g$}
            \State Refresh $D$, groups, and $\mathbf{W}$ using \paperref{Eqs.}{eq:cosine_sim}--\ref{eq:soft_weights}
            \State Use the refreshed weights from the next iteration
        \EndIf
    \EndIf
\EndFor
\end{algorithmic}
\end{algorithm}

\subsection{Alternative Design Axes}\label{sec:alt_axes}
The following comparisons vary the distance construction, gradient processing, or architecture. Full variant equations and implementation settings are collected in~\ref{app:methods}.

\subsubsection{Gradient-only grouping}
The Direct variant uses only gradient relationships when refreshing groups; its initial partition still comes from static EG. It uses smoothed gradient similarities and balanced $k$-medoids to form groups; its soft version uses a kernel based on those smoothed similarities to calculate group affinities.

\subsubsection{Geodesic distances (GeoEG)}\label{sec:exp_gradient}
GeoEG replaces the unweighted shortest-path structural distances used by static EG with geometry-weighted shortest-path distances. We evaluate fixed groups formed from the resulting FGW matrix and dynamic configurations initialized with it, labelled GeoEG Initialization in the tables.

\subsubsection{Gradient processing}
We evaluate symmetric all-pairs PCGrad projections, including normalized-gradient and inter-group variants, within the EG update framework~\citep{yu2020pcgrad}. These are implementation-specific variants of gradient projection.

\subsubsection{Architecture}
The reported architecture comparisons use per-robot output heads or an actor mixture of experts. The tested MoE is inspired by MOORE and differs from the original implementation~\citep{hendawy2024moore}; its result applies to the implemented variant.

\subsubsection{Ablations of DEG}\label{sec:deg_ablations}
Two additional variants change one mechanism of DEG. DEG-Freeze disables further regrouping after the first successful refresh, retaining the resulting soft weights and the sequential group updates. DEG-Collapse continues refreshing but applies the same mean of the group gradients at every sequential step, removing group-specific updates. Their exact definitions are provided in~\ref{app:methods}.

\section{Experimental Setup}\label{sec:setup}

\paragraph{Training setup}
All configurations use the IQL objectives described in \paperref{Section}{sec:backbone}. Main grouped configurations use $M=4$ groups; we also evaluate $M=7$ in the cross-dataset comparison. The default URMA-based architecture, actor and critic variants, and numerical training settings are summarized in~\ref{app:training}; variant definitions are given in~\ref{app:methods}. 

\paragraph{Evaluation protocol}
Each configuration uses five training seeds (0--4). At the final evaluation after 100K training iterations, we evaluate each robot for 30 episodes per seed using deterministic policy-mean actions. For each seed, we first average episodic return within each robot and then average equally across all 16 robots. We report the mean and population standard deviation of these five seed-level averages. A robot is considered ``rescued'' within a seed if its episode-averaged return is $\geq 10$; reported mean rescue counts average the resulting counts across seeds.

On the primary 70\% Suboptimal Forward dataset, we compare 31 configurations covering dynamic grouping, gradient projection, architecture, geodesic distances, soft assignments, and inter-group PCGrad. The two DEG ablations and comparisons of selected configurations on Expert and Replay data are reported separately in \paperref{Section}{sec:results}.

The released EG implementation specifies different evaluation-episode counts in its IQL+EG experiment script (one episode) and default configuration (10 episodes).\footnote{Released EG implementation: \url{https://github.com/haruki-abe/cross_embodiment_offline_rl/tree/552ea9b}.} The EG paper does not identify the count used for its published results. We therefore state our protocol explicitly and interpret comparisons with the published returns of Abe et al.~\citep{abe2024eg} in light of this uncertainty. All within-study method comparisons, including the static-EG baseline, use the 30-episode protocol above.

\section{Results}\label{sec:results}

We report 31 configurations across 16 robots on the 70\% Suboptimal Forward @100K benchmark (\paperref{Table}{tab:main_results}; full configuration results appear in~\ref{app:full_results}, \paperref{Table}{tab:appendix_full_results}, and per-robot results in~\ref{app:per_robot}). We first compare DEG with static EG, then present results for the other grouping, gradient-processing, and architecture configurations. Finally, we compare selected configurations across the three dataset variants.

\begin{table}[!ht]
\centering
\small
\renewcommand{\baselinestretch}{1}\selectfont  
\setlength{\tabcolsep}{2.5pt}
\renewcommand{\arraystretch}{0.85}
\begin{tabular}{lccr}
\toprule
\textbf{Method} & \textbf{Mean} & \textbf{Std} & \textbf{$\Delta$} \\
\midrule
\multicolumn{4}{l}{\textit{Baselines}} \\
\quad Uniform (no groups) & 41.89 & 2.15 & $-6.39$ \\
\quad Group-Uniform (1 group/step) & 31.62 & 7.98 & $-16.66$ \\
\midrule
\multicolumn{4}{l}{\textit{Static grouping}} \\
\quad EG Static & 48.28 & 5.88 & -- \\
\midrule
\multicolumn{4}{l}{\textit{Dynamic grouping}} \\
\quad \textbf{EG Dynamic Soft (DEG)} & \textbf{52.79} & \textbf{2.13} & \textbf{+4.51} \\
\quad EG Dynamic Direct + Inter-PCGrad & 50.95 & 2.76 & $+2.67$ \\
\quad EG Dynamic Direct & 50.92 & 3.32 & $+2.64$ \\
\quad EG Dynamic (FGW) & 49.81 & 5.57 & +1.53 \\
\quad EG Dynamic Direct Soft & 44.79 & 4.81 & -3.49 \\
\midrule
\multicolumn{4}{l}{\textit{Gradient projection}} \\
\quad EG Dynamic Direct + PCGrad Norm & 50.77 & 2.23 & $+2.49$ \\
\quad EG FGW Soft + PCGrad & 50.35 & 3.17 & +2.07 \\
\midrule
\multicolumn{4}{l}{\textit{Architecture variants}} \\
\quad EG Static + Per-Robot Heads & 50.05 & 1.80 & +1.77 \\
\quad EG Static + Per-Robot Actor Heads & 49.46 & 1.75 & +1.18 \\
\quad EG Static + MoE & 30.66 & 2.94 & -17.62 \\
\bottomrule
\end{tabular}
\caption{Main results on 70\% Suboptimal Forward @100K. $\Delta$ is the difference from EG Static. Std across 5 seeds. Full reported comparison in \paperref{Table}{tab:appendix_full_results}.}
\label{tab:main_results}
\end{table}

\subsection{Primary-Dataset Results and Exploratory Configurations}

\paragraph{DEG and static EG}
The learning curves in \paperref{Appendix Figure}{fig:app_sub70_curves} show DEG attaining higher mean returns than static EG late in training; DEG also has the highest final mean among the selected methods. \paperref{Figure}{fig:entropy} separately summarizes DEG's soft-weight entropy over training.
DEG achieves a mean return of 52.79, compared with 48.28 for static EG, a difference of 4.51 points. The standard deviations across five seeds are 2.13 and 5.88, respectively. Per-robot changes are not uniform: talos, nao\_v5, a1, and h1 each improve by more than 10 points, while other robots change less or decline (\paperref{Table}{tab:appendix_per_robot}). A possible explanation is that DEG changes how robot gradients are combined, so the shared actor updates may align more closely with the learning objectives of some robots and less closely with those of others.

\paragraph{Exploratory hard- and soft-grouping configurations}
We implemented FGW-based and gradient-only variants with both hard and soft assignments to compare these choices in the evaluated setting. Under FGW regrouping, DEG (the soft configuration) reaches 52.79, compared with 49.81 for the hard configuration. Under gradient-only regrouping, soft assignment reaches 44.79, compared with 50.92 for hard assignment. Thus, the two variants produce different outcomes. The two constructions use FGW with Ward clustering and gradient-only distances with balanced $k$-medoids, respectively. \paperref{Figure}{fig:group_sizes} and the soft-weight entropy analysis in~\ref{app:entropy} report group sizes and entropy. The rescue counts and DEG group assignments over training are shown in~\ref{app:rescued}.

\begin{figure}[!ht]
\centering
\includegraphics[width=0.8\linewidth]{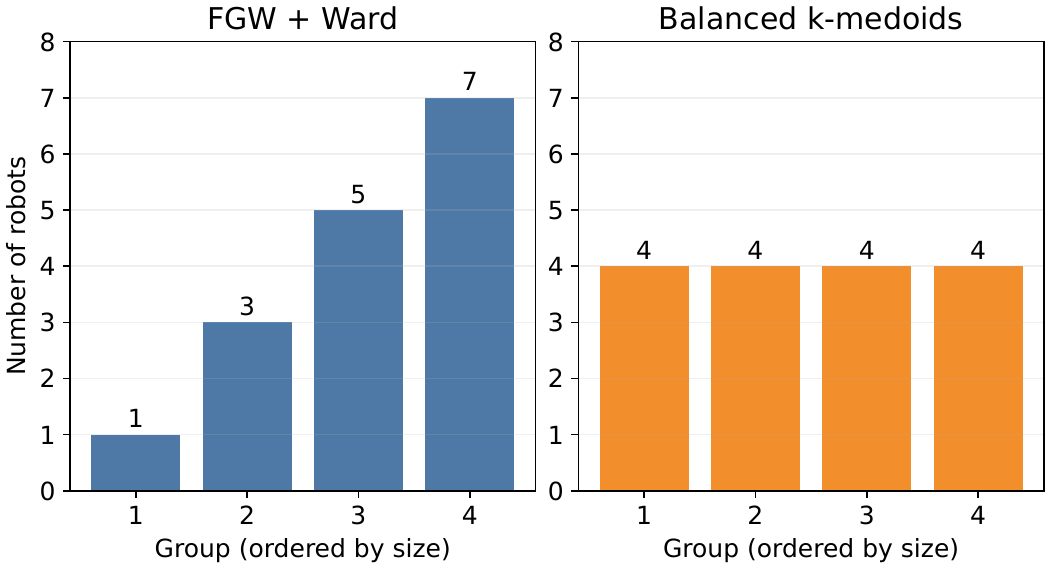}
\caption{Group sizes produced by the evaluated clustering configurations. The displayed FGW+Ward partition has unequal group sizes (1, 3, 5, and 7 robots), while the gradient-only $k$-medoids partition has four robots per group (4--4--4--4).}
\label{fig:group_sizes}
\end{figure}

\paragraph{Geodesic-distance configurations}
GeoEG Static reaches a mean of 45.92, compared with 48.28 for EG Static; on talos, their returns are 36.3 and 89.3, respectively.

\begin{figure}[!ht]
\centering
\includegraphics[width=0.8\linewidth]{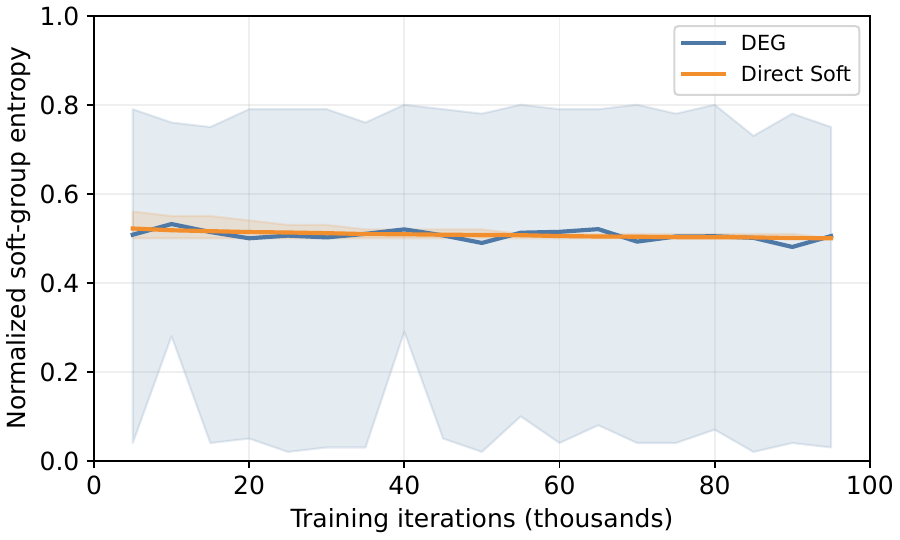}
\caption{Normalized soft-group entropy over training. Lines show the mean across 4 groups and 5 seeds; shaded bands show the minimum and maximum group entropies across groups and seeds at each refresh. Individual group entropies range from 0.02 to 0.80 for DEG and from 0.50 to 0.56 for EG Dynamic Direct Soft.}
\label{fig:entropy}
\end{figure}

\subsection{Ablation Variants of DEG}

We evaluate the DEG-Freeze and DEG-Collapse ablations defined in \hyperref[sec:deg_ablations]{\S\ref*{sec:deg_ablations}} on the primary dataset with five seeds under the protocol in \hyperref[sec:setup]{\S\ref*{sec:setup}} (\paperref{Table}{tab:ablation}).

\begin{table}[!ht]
\centering
\small
\renewcommand{\baselinestretch}{1}\selectfont  
\setlength{\tabcolsep}{2.5pt}
\renewcommand{\arraystretch}{0.85}
\begin{tabular}{lccr}
\toprule
\textbf{Variant} & \textbf{Mean} & \textbf{Std} & \textbf{$\Delta$} \\
\midrule
\quad EG Static & 48.28 & 5.88 & -- \\
\quad \textbf{EG Dynamic Soft (DEG)} & \textbf{52.79} & \textbf{2.13} & \textbf{+4.51} \\
\midrule
\quad DEG-Freeze & 49.81 & 4.55 & +1.53 \\
\quad DEG-Collapse & 49.79 & 2.89 & +1.51 \\
\bottomrule
\end{tabular}
\caption{Ablation variants of DEG on 70\% Suboptimal Forward @100K, five seeds each, evaluated under the protocol of \hyperref[sec:setup]{\S\ref*{sec:setup}}; std across seeds as in \paperref{Table}{tab:main_results}. $\Delta$ is the difference from EG Static. Both variants share DEG's initial grouping and first successful refresh (\hyperref[sec:deg_ablations]{\S\ref*{sec:deg_ablations}}).}
\label{tab:ablation}
\end{table}

DEG-Freeze and DEG-Collapse reach $49.81\pm4.55$ and $49.79\pm2.89$, respectively, compared with $52.79\pm2.13$ for DEG (mean $\pm$ population standard deviation across five seeds). Both ablations remain above static EG's mean of 48.28, but their means are about 3 points below DEG's. Their lowest seed returns (40.88 and 45.66) are below DEG's lowest (50.16). DEG-Freeze has one seed below this minimum, and DEG-Collapse has two.

\subsection{Other Evaluated Configurations}

In addition to the grouping variants, we tested gradient-projection and architecture configurations. Their aggregate and per-robot results are reported below and in \paperref{Figure}{fig:per_robot}.

\paragraph{Gradient-projection variants}
EG Dynamic Direct + PCGrad Norm and EG FGW Soft + PCGrad reach mean returns of 50.77 and 50.35, respectively, compared with 52.79 for DEG. For EG Dynamic Direct + PCGrad Norm, the per-robot table shows a 21.6-point increase for talos and a 7.6-point decrease for go1 relative to static EG (\paperref{Table}{tab:appendix_per_robot}). Both of these configurations have higher mean returns than static EG and lower mean returns than DEG.

\paragraph{Architecture variants}
EG Static + Per-Robot Heads reaches 50.05, with per-robot changes that include increases for talos and h1 and a decrease for g1 relative to static EG (\paperref{Table}{tab:appendix_per_robot}). Actor-only and critic-only heads reach 49.46 and 48.47, respectively. The tested MoE variant reaches 30.66. Under the rescue threshold of \hyperref[sec:setup]{\S\ref*{sec:setup}}, DEG averages 13.8 rescued robots and the tested MoE variant 9.0 (\paperref{Figure}{fig:rescued_robots}).

\paperref{Figure}{fig:per_robot} shows per-robot returns for selected methods. DEG has the highest aggregate mean among the methods shown, while the per-robot returns vary across methods.

\begin{figure}[t]
\centering
\includegraphics[width=0.98\textwidth]{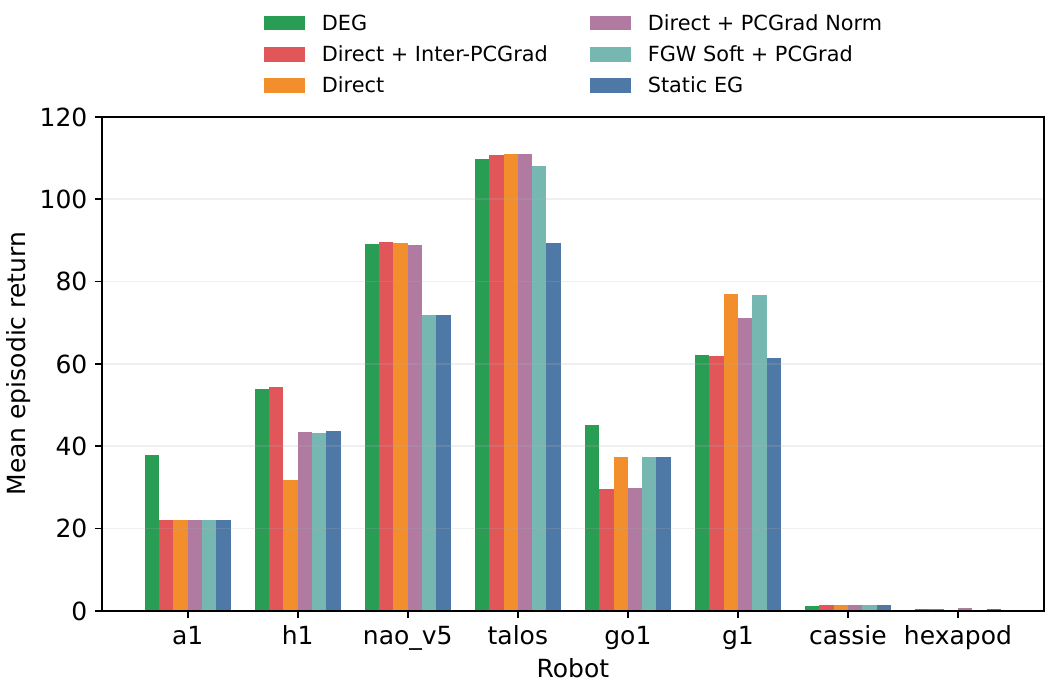}
\caption{Per-robot mean episodic return for six selected methods on 70\% Suboptimal data. The x-axis shows eight robots: unitree\_a1, unitree\_h1, nao\_v5, talos, unitree\_go1, unitree\_g1, cassie, and hexapod; full results for all 16 robots are in \paperref{Table}{tab:appendix_per_robot}. \paperref{Table}{tab:appendix_full_results} reports seed standard deviations for the aggregate returns averaged across all 16 robots. Cassie and hexapod have returns below 2 for the evaluated methods.}
\label{fig:per_robot}
\end{figure}

\subsection{Results Across Dataset Variants}

\paperref{Table}{tab:data_quality} compares selected configurations across suboptimal, expert, and replay data to examine how their relative performance changes across conditions.

\begin{table}[!ht]
\centering
\small
\renewcommand{\baselinestretch}{1}\selectfont  
\setlength{\tabcolsep}{3pt}
\begin{tabular}{lccc}
\toprule
\textbf{Method} & \textbf{Sub70} & \textbf{Expert} & \textbf{Replay} \\
\midrule
EG Dynamic Soft (DEG, $M{=}4$) & $52.79\pm2.13$ & $63.42\pm0.09$ & $53.79\pm1.95$ \\
EG Dynamic Soft ($M{=}7$) & $50.65\pm4.42$ & $63.31\pm0.24$ & $55.13\pm0.89$ \\
EG FGW Soft + PCGrad & $50.35\pm3.17$ & $63.50\pm0.11$ & $55.33\pm0.84$ \\
EG Dynamic Direct + PCGrad Norm & $50.77\pm2.23$ & $63.40\pm0.12$ & $53.68\pm1.17$ \\
EG Dynamic Direct + Inter-PCGrad & $50.95\pm2.76$ & $62.40\pm1.85$ & $53.79\pm0.89$ \\
EG Dynamic Direct & $50.92\pm3.32$ & $63.47\pm0.11$ & $55.12\pm1.12$ \\
EG Static & $48.28\pm5.88$ & $63.53\pm0.07$ & $54.57\pm1.04$ \\
\bottomrule
\end{tabular}
\caption{Mean episodic return $\pm$ population standard deviation across five seeds on the three Forward @100K dataset variants.}
\label{tab:data_quality}
\end{table}

\paragraph{Expert dataset}
On expert data, the reported means for the listed methods range from 62.40 to 63.53; static EG reaches $63.53\pm0.07$ and DEG reaches $63.42\pm0.09$ (\paperref{Table}{tab:data_quality}). DEG's mean is 0.11 points below static EG's.

\paragraph{Replay dataset}
On replay data, DEG reaches $53.79\pm1.95$, compared with $54.57\pm1.04$ for static EG, a mean-return difference of 0.78 points; EG FGW Soft + PCGrad reaches $55.33\pm0.84$ (\paperref{Table}{tab:data_quality}). Cassie and hexapod returns range from 72 to 82 for the reported replay configurations (\paperref{Table}{tab:appendix_per_robot_replay}), compared with values below 2 on suboptimal data.

\paragraph{Group-count configurations}
For DEG with $M{=}4$ and $M{=}7$, the reported means are 52.79 and 50.65 on suboptimal data, 63.42 and 63.31 on expert data, and 53.79 and 55.13 on replay data (\paperref{Table}{tab:data_quality}). The higher-performing group-count configuration differs between the suboptimal and replay datasets; the expert means are similar (\paperref{Figure}{fig:M_sensitivity}).

\begin{figure}[!ht]
\centering
\includegraphics[width=0.7\linewidth]{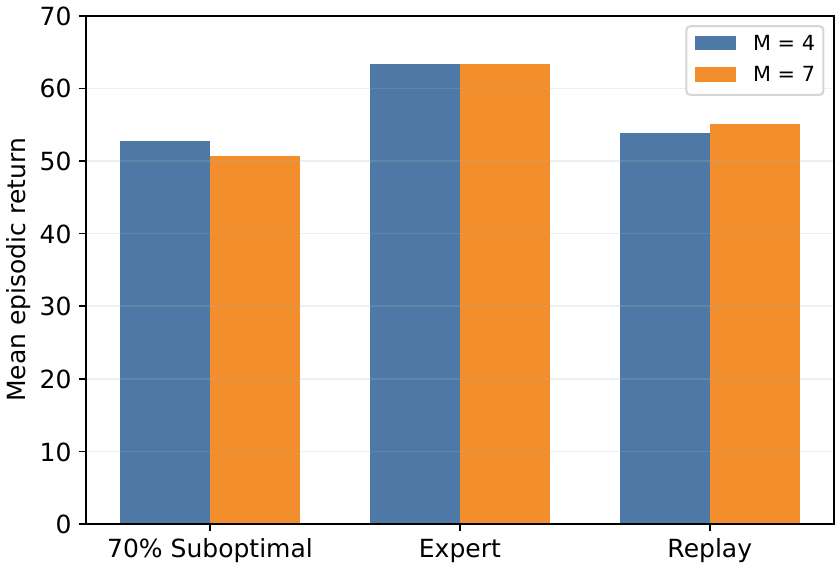}
\caption{Reported means for DEG with $M{=}4$ and $M{=}7$ across three dataset variants. The relative ordering differs between suboptimal and replay data; the expert means are similar.}
\label{fig:M_sensitivity}
\end{figure}

The higher mean return and lower seed variability on 70\% Suboptimal Forward, almost identical Expert Forward means, and absence of improvement on Replay Forward suggest that DEG may be particularly useful when learning from suboptimal demonstrations. Computational cost is analyzed in~\ref{app:cost}.

\section{Limitations}\label{sec:discussion}

Only one suboptimal-data mixture was evaluated, so broader evaluation is needed to determine when DEG benefits learning from suboptimal data. The dataset-dependent ordering should also be interpreted alongside seed variability (\paperref{Section}{sec:results}).

Evaluation covers one simulated locomotion task and the reported datasets, without testing on real robots or establishing performance on other tasks or platforms. Most configurations use a shared actor with a three-hidden-layer MLP following the URMA encoders, and expert-data results alone do not establish an architectural ceiling.

\section{Conclusion}

We presented DEG for cross-embodiment offline reinforcement learning. DEG combines morphology and current actor-gradient information to periodically refresh robot groups and compute soft contributions to sequential group updates. It retains the shared IQL backbone and grouped actor-update framework of EG. We compared 31 configurations covering grouping, gradient processing, and policy architecture on a benchmark of 16 simulated robots.

On 70\% Suboptimal Forward, DEG achieves a mean return of 52.79, compared with 48.28 for static EG, an improvement of 4.51 points under the common 30-episode evaluation protocol. Across five seeds, its standard deviation is 2.13, compared with 5.88 for static EG. These results demonstrate improved learning from suboptimal data in the evaluated setting.


\section*{CRediT authorship contribution statement}
\textbf{AmirAli Fakhari Zavareh:} Conceptualization, Methodology, Software, Investigation, Data curation, Writing -- original draft, Writing -- review \& editing.
\textbf{Jianming Ma:} Methodology, Investigation, Writing -- review \& editing.
\textbf{Zhenzhe Sun:} Writing -- review \& editing (proposed, pending author confirmation).
\textbf{Yue Gao:} Conceptualization, Supervision, Writing -- review \& editing.

\section*{Declaration of competing interest}
The authors declare that they have no known competing financial interests or personal relationships that could have appeared to influence the work reported in this paper.

\section*{Funding}
This research did not receive any specific grant from funding agencies in the public, commercial, or not-for-profit sectors.

\section*{Data availability}
Offline locomotion data:
\par\noindent{\footnotesize\url{https://huggingface.co/datasets/haruki-abe/cross_embodiment_offline_rl}}.

\smallskip
The \href{https://rl2-theory.github.io/deg-project-page/}{DEG project page} provides results and visualizations. The training code and experiment scripts are maintained in a private repository.


\appendix

\counterwithin*{table}{section}
\counterwithin*{figure}{section}

\clearpage
\section*{Appendix contents}
\begingroup
\small\singlespacing
\renewcommand{\arraystretch}{1.15}
\noindent\begin{tabular}{@{}>{\raggedright\arraybackslash}p{\dimexpr\linewidth-4em\relax}@{\hspace{1em}}r@{}}
\textbf{Appendix} & \textbf{Page} \\[6pt]
\hyperref[app:methods]{\textbf{\ref*{app:methods}\quad\nameref*{app:methods}}} & \hyperref[app:methods]{\pageref*{app:methods}} \\[5pt]
\hspace*{1em}\hyperref[app:a_problem]{\ref*{app:a_problem}\quad\nameref*{app:a_problem}} & \hyperref[app:a_problem]{\pageref*{app:a_problem}} \\[5pt]
\hspace*{1em}\hyperref[app:a_backbone]{\ref*{app:a_backbone}\quad\nameref*{app:a_backbone}} & \hyperref[app:a_backbone]{\pageref*{app:a_backbone}} \\[5pt]
\hspace*{1em}\hyperref[app:a_static_eg]{\ref*{app:a_static_eg}\quad\nameref*{app:a_static_eg}} & \hyperref[app:a_static_eg]{\pageref*{app:a_static_eg}} \\[5pt]
\hspace*{1em}\hyperref[app:a_deg]{\ref*{app:a_deg}\quad\nameref*{app:a_deg}} & \hyperref[app:a_deg]{\pageref*{app:a_deg}} \\[5pt]
\hspace*{1em}\hyperref[app:a_variants]{\ref*{app:a_variants}\quad\nameref*{app:a_variants}} & \hyperref[app:a_variants]{\pageref*{app:a_variants}} \\[5pt]
\hspace*{1em}\hyperref[app:a_ablations]{\ref*{app:a_ablations}\quad\nameref*{app:a_ablations}} & \hyperref[app:a_ablations]{\pageref*{app:a_ablations}} \\[5pt]
\hyperref[app:entropy]{\textbf{\ref*{app:entropy}\quad\nameref*{app:entropy}}} & \hyperref[app:entropy]{\pageref*{app:entropy}} \\[5pt]
\hyperref[app:full_results]{\textbf{\ref*{app:full_results}\quad\nameref*{app:full_results}}} & \hyperref[app:full_results]{\pageref*{app:full_results}} \\[5pt]
\hyperref[app:additional_curves]{\textbf{\ref*{app:additional_curves}\quad\nameref*{app:additional_curves}}} & \hyperref[app:additional_curves]{\pageref*{app:additional_curves}} \\[5pt]
\hyperref[app:per_robot]{\textbf{\ref*{app:per_robot}\quad\nameref*{app:per_robot}}} & \hyperref[app:per_robot]{\pageref*{app:per_robot}} \\[5pt]
\hyperref[app:per_robot_datasets]{\textbf{\ref*{app:per_robot_datasets}\quad\nameref*{app:per_robot_datasets}}} & \hyperref[app:per_robot_datasets]{\pageref*{app:per_robot_datasets}} \\[5pt]
\hyperref[app:rescued]{\textbf{\ref*{app:rescued}\quad\nameref*{app:rescued}}} & \hyperref[app:rescued]{\pageref*{app:rescued}} \\[5pt]
\hyperref[app:training]{\textbf{\ref*{app:training}\quad\nameref*{app:training}}} & \hyperref[app:training]{\pageref*{app:training}} \\[5pt]
\hyperref[app:cost]{\textbf{\ref*{app:cost}\quad\nameref*{app:cost}}} & \hyperref[app:cost]{\pageref*{app:cost}} \\[5pt]
\hspace*{1em}\hyperref[app:i_parameters]{\ref*{app:i_parameters}\quad\nameref*{app:i_parameters}} & \hyperref[app:i_parameters]{\pageref*{app:i_parameters}} \\[5pt]
\hspace*{1em}\hyperref[app:i_runtime]{\ref*{app:i_runtime}\quad\nameref*{app:i_runtime}} & \hyperref[app:i_runtime]{\pageref*{app:i_runtime}} \\[5pt]
\end{tabular}
\endgroup

\clearpage
\section{Detailed Method Formulation and Implementation}
\label{app:methods}

This appendix gives the offline-data definitions, IQL losses, static-EG updates, DEG implementation, comparisons, and ablations, in that order. It supplements \paperref{Sections}{sec:backbone} and~\ref{sec:exp_grouping}; numerical settings are collected in \paperref{Table}{tab:hyperparameters}.

\subsection{Problem Setting and Offline Data}
\label{app:a_problem}

\subsubsection{Cross-Embodiment Offline RL}

Let $\mathcal{T}=\{1,\dots,N\}$ denote a set of robot embodiments. Each embodiment $\tau\in\mathcal{T}$ induces an MDP
\begin{equation}
\mathcal{M}_{\tau} = (\mathcal{S}_{\tau},\mathcal{A}_{\tau},P_{\tau},r_{\tau},\rho_{\tau},\gamma),
\end{equation}
where $\mathcal{S}_{\tau}$ and $\mathcal{A}_{\tau}$ are embodiment-specific state and action spaces, $P_{\tau}(s'|s,a)$ is the transition dynamics, $r_{\tau}(s,a)$ is the reward function, $\rho_{\tau}$ is the initial-state distribution, and $\gamma\in(0,1)$ is the discount factor. The spaces differ because robots may have different numbers of joints, limbs, feet, and actuators. For each embodiment, we assume access only to a static offline dataset
\begin{equation}
\mathcal{D}_{\tau}=\{(s_t,a_t,r_t,s_{t+1},d_t)\}_{t=1}^{|\mathcal{D}_{\tau}|},
\end{equation}
with $d_t$ marking episode termination. The datasets are collected by behavior policies that are not queried during training.

The shared objective in \paperref{Eq.}{eq:ce_objective} uses uniform $p(\tau)$. All policy and value updates use the offline datasets, whose disjoint union is $\mathcal{D}=\bigsqcup_{\tau\in\mathcal{T}}\mathcal{D}_{\tau}$.

\subsubsection{Task and Offline Datasets}

Each robot is tasked with tracking a forward command of $1$\,m/s on flat terrain in the MuJoCo physics simulator~\citep{todorov2012mujoco}. The dense reward follows~\citet{bohlinger2024urma}: it rewards commanded planar- and yaw-velocity tracking and provides penalties for vertical and angular motion, posture, joint limits, joint velocity and acceleration, torque, action rate, trunk height, collisions, and gait asymmetry, together with a signed airtime term. Individual terms are weighted by robot-specific coefficients, some of which are zero. The total reward is clipped below at zero.

The dataset suite and download link are given in \paperref{Section}{sec:backbone}. The following details clarify data construction and the naming correspondence with~\citet{abe2024eg}:
\begin{itemize}[leftmargin=*,nosep]
    \item \textbf{Expert Forward:} 1M transitions collected by rolling out a fully converged Proximal Policy Optimization (PPO) policy (expert data).
    \item \textbf{Replay Forward} (called Expert Replay by Abe et al.): 1M transitions uniformly subsampled from PPO training data collected from the start of training through expert-level performance (mixed quality).
    \item \textbf{70\% Suboptimal Forward} (called 70\% Suboptimal Replay by Abe et al.): 70\% of transitions sampled from the early, suboptimal phase of PPO training and 30\% from the later, expert-like phase.
\end{itemize}
For the locally available Suboptimal and Replay Forward variants, the 1M-transitions-per-robot counts stated in \paperref{Section}{sec:backbone} are specified in the metadata. We use the released Expert data and rely on the EG paper and repository for its construction and split definitions.

\subsection{Shared Backbone and IQL Objectives}
\label{app:a_backbone}

\subsubsection{Backbone Architecture and Parameter Sharing}

The URMA observation decomposition~\citep{bohlinger2024urma} accommodates different observation and action dimensions. Each observation is split into a fixed-dimensional general input $o_g$ and variable-length joint and foot inputs. Let $J(\tau)$ and $F(\tau)$ be the joint and foot index sets of embodiment $\tau$. For each joint $j\in J(\tau)$ and foot $f\in F(\tau)$, the corresponding input consists of its descriptor ($d_j$ or $d_f$) and state observation ($o_j$ or $o_f$). Descriptor-conditioned attention maps each set to a fixed-dimensional latent vector. Concatenating both latents with $o_g$ gives the shared input representation $z_{\mathrm{shared}}(s,\tau)$.

The default model has four independently parameterized online networks: actor $\pi_{\theta}$, value function $V_{\psi}$, and twin Q-functions $Q_{\phi_1}$ and $Q_{\phi_2}$. The twin critics have identical architectures and separate parameters; their target copies $\bar{Q}_{\phi_1}$ and $\bar{Q}_{\phi_2}$ supply the expectile value update.

In the default configuration, each network's parameters are shared across all $N$ robots; robot conditioning is provided by the descriptors and observation inputs, including robot-specific features in $o_g$, rather than by per-robot parameters. The per-robot-head and MoE variants in \paperref{Section}{sec:alt_axes} add robot-indexed parameters.

\subsubsection{IQL Objectives}

The IQL critic losses~\citep{kostrikov2022iql} use dataset actions. For embodiment $\tau$, the value function minimizes the expectile loss
\begin{equation}
\mathcal{L}^{V}_{\tau}(\psi)
= \mathbb{E}_{(s,a)\sim\mathcal{D}_{\tau}}
\left[
L_2^{\omega}\left(Q_{\bar{\phi}}(s,a,\tau)-V_{\psi}(s,\tau)\right)
\right],
\label{eq:iql_v}
\end{equation}
where $L_2^{\omega}(u)=|\omega-\mathds{1}(u<0)|u^2$ is the asymmetric squared loss for scalar residual $u$, with expectile $\omega$. The indicator $\mathds{1}(u<0)$ is one when $u<0$ and zero otherwise; $Q_{\bar{\phi}}$ denotes the minimum of the two target critic outputs. The action-value functions are trained with the temporal-difference objective
\begin{equation}
\mathcal{L}^{Q}_{\tau}(\phi_1,\phi_2)
= \mathbb{E}_{(s,a,r,s',d)\sim\mathcal{D}_{\tau}}
\left[
\sum_{k=1}^{2}\left(Q_{\phi_k}(s,a,\tau)-y\right)^2
\right],
\label{eq:iql_q}
\end{equation}
with target $y=r+\gamma(1-d)V_{\psi}(s',\tau)$. The actor objective, advantage weights, and valid-action-dimension convention are specified in \paperref{Section}{sec:backbone}.
The critic losses are computed per embodiment and averaged over all robots at each iteration, while the actor update is organized at the group level (\paperref{Section}{sec:eg_training}).

\subsection{Static Embodiment Grouping}
\label{app:a_static_eg}

Static EG~\citep{abe2024eg} defines the fixed groups and sequential actor-update framework used as the baseline.

\subsubsection{Morphology Graphs, FGW Dissimilarities, and Fixed Groups}

Following the morphology-graph representation used by EG~\citep{abe2024eg}, each robot $i$ is represented as an undirected kinematic graph $\mathcal{G}_i = (\mathcal{V}_i, \mathcal{E}_i)$. Nodes represent the robot's joints and feet, with a trunk node where required by the kinematic tree. Each node has a feature vector $\mathbf{x}_p \in \mathbb{R}^d$ derived from the robot description; these descriptors include physical and actuator attributes such as joint limits, torque and velocity limits, and link geometry. Missing descriptor entries are zero-padded. Edges represent parent--child connections derived from the MuJoCo model tree.

EG uses Fused Gromov--Wasserstein (FGW) optimal transport to compute pairwise robot dissimilarities~\citep{vayer2019fgw}. Let $\mu_i$ and $\mu_j$ be uniform probability distributions over the nodes of graphs $\mathcal{G}_i$ and $\mathcal{G}_j$, respectively. For node pairs $(p,r)\in\mathcal{V}_i^2$ and $(q,s)\in\mathcal{V}_j^2$, the FGW objective is
\begin{equation}
\begin{aligned}
d_{\text{FGW}}(\mathcal{G}_i,\mathcal{G}_j) \triangleq
\min_{\mathbf{T}\in\Pi(\mu_i,\mu_j)} \Bigg[ & (1-\alpha)\sum_{p,q}\mathbf{T}_{pq}\,\|\mathbf{x}_p - \mathbf{x}_q\|^2 \\
&+ \alpha\sum_{p,q,r,s}\mathbf{T}_{pq}\mathbf{T}_{rs}\,(\mathbf{C}_i(p,r)-\mathbf{C}_j(q,s))^2 \Bigg],
\end{aligned}
\label{eq:fgw}
\end{equation}
where $\Pi(\mu_i,\mu_j)$ is the set of nonnegative transport plans whose row and column marginals are $\mu_i$ and $\mu_j$, and $\mathbf{T}_{pq}$ is the mass coupling node $p$ to node $q$. The matrices $\mathbf{C}_i$ and $\mathbf{C}_j$ contain unweighted all-pairs shortest-path lengths within the respective graphs.\footnote{Released EG distance generator: \url{https://github.com/haruki-abe/cross_embodiment_offline_rl/blob/552ea9b/experiments/create_distance_matrix.py}.} The coefficient $\alpha\in[0,1]$ weights structural alignment relative to node-feature alignment; the implementation uses $\alpha=0.5$.

We numerically optimize this objective using POT's conditional-gradient solver with an Armijo line search.\footnote{\url{https://pythonot.github.io/gen_modules/ot.gromov.html\#ot.gromov.fused_gromov_wasserstein}} The implementation passes \texttt{epsilon=0.01}, but this keyword does not introduce entropy regularization in the called \texttt{fused\_gromov\_wasserstein} solver. We use the final objective value returned by the solver. We call the resulting value an FGW dissimilarity, since the objective is not asserted here to satisfy all metric axioms.

The resulting $N\times N$ pairwise FGW dissimilarity matrix is divided by its maximum value, scaling its entries to $[0,1]$, and is supplied to Ward hierarchical clustering~\citep{ward1963hierarchical} to obtain $M$ groups. These groups remain fixed during training in Static EG.

The implementation applies Ward linkage directly to the condensed FGW dissimilarities; because FGW dissimilarities are not guaranteed to be Euclidean, the usual variance-minimization interpretation of Ward linkage is not guaranteed in this application. The FGW-based configurations share this graph representation, FGW objective, and Ward clustering procedure; the dynamic FGW variants modify the node features and structural matrices before computing FGW (\paperref{Section}{sec:exp_grouping}).

\subsubsection{Gradient Aggregation and Sequential Group Updates}

At each iteration, one minibatch is sampled per embodiment. Its actor gradient is $\bm{g}_{\tau}=\nabla_{\theta}\mathcal{L}^{\pi}_{\tau}(\theta)$ at the shared actor parameters $\theta$. A joint update would aggregate these gradients as
\begin{equation}
\bm{g}_{\mathrm{joint}} = \sum_{\tau\in\mathcal{T}} p(\tau)\bm{g}_{\tau},
\end{equation}
where $p(\tau)$ is the weight assigned to embodiment $\tau$ (uniform across embodiments in our experiments). The implementation passes the aggregate to Adam rather than applying a plain stochastic-gradient-descent step. For two distinct embodiments $\tau$ and $\tau'$, a negative inner product $\bm{g}_{\tau}^{\top}\bm{g}_{\tau'}<0$ indicates opposing gradient directions. Their contributions can partially cancel in a joint update, which can contribute to negative transfer.

For $M$ morphology-based groups $\{G_m\}_{m=1}^{M}$~\citep{abe2024eg}, the value and twin-Q losses are averaged over all $N$ embodiment minibatches. Writing $\psi$ for the value-network parameters and $\phi_1,\phi_2$ for the two Q-network parameter sets, the two objectives are
\begin{equation}
\mathcal{L}^{Q}(\phi_1,\phi_2) = \frac{1}{N}\sum_{\tau\in\mathcal{T}}\mathcal{L}^{Q}_{\tau}(\phi_1,\phi_2),
\qquad
\mathcal{L}^{V}(\psi) = \frac{1}{N}\sum_{\tau\in\mathcal{T}}\mathcal{L}^{V}_{\tau}(\psi).
\end{equation}
Here $\mathcal{L}^{Q}_{\tau}$ is the twin-Q loss and $\mathcal{L}^{V}_{\tau}$ is the value loss for embodiment $\tau$, as defined in \paperref{Eqs.}{eq:iql_q} and~\ref{eq:iql_v}. The implementation applies each objective using its corresponding optimizer.

The actor is then updated sequentially at the group level. For each group $G_m$, the group gradient is the unweighted mean of the member embodiments' actor gradients,
\begin{equation}
\bar{\bm{g}}_{m} = \frac{1}{|G_m|}\sum_{\tau\in G_m} \bm{g}_{\tau},
\label{eq:eg_prelim_update}
\end{equation}
where $|G_m|$ is the number of embodiments in group $m$. The implementation passes each $\bar{\bm{g}}_m$ to the shared Adam optimizer in sequence, carrying its optimizer state between group updates. The groups are visited in the same pseudorandomly generated order at every iteration. The policy, value, and Q networks retain their shared representations across these group steps.

\subsection{Dynamic Soft Embodiment Grouping}
\label{app:a_deg}

DEG replaces fixed actor grouping with periodically refreshed groups and soft contributions; the networks and critic losses remain as defined above.

\subsubsection{Gradient Information for Regrouping}

The pairwise cosine similarity is defined in \paperref{Eq.}{eq:cosine_sim}. The gradients are evaluated on one minibatch per robot at common current policy parameters; the diagonal is set explicitly to $S_{ii}=1$. The cosine denominator uses $\epsilon_{\mathrm{cos}}=10^{-8}$. Its robot-wise mean and change supply the augmented graph features in \paperref{Eq.}{eq:aug_features}, while pairwise dissimilarities enter the blend in \paperref{Eq.}{eq:blend}. The corresponding feature scales, mixing coefficient, and numerical guards are listed in \paperref{Table}{tab:hyperparameters}.

\subsubsection{FGW-Based Regrouping}

For every dynamic FGW configuration, the implementation constructs kinematic graphs with geometry-weighted edges, weighting each edge $(u,v)$ by the default-pose Euclidean distance
\begin{equation}
w_{uv} = \max(\|\mathrm{pos}(u) - \mathrm{pos}(v)\|_2,\; \varepsilon_{\text{min}}),
\label{eq:geo_weight}
\end{equation}
where $\mathrm{pos}(u)$ is the default-pose position of the body associated with a joint or trunk node, or of the geometry associated with a foot node. All-pairs weighted Dijkstra distances~\citep{dijkstra1959note} form the per-robot structural matrices $\mathbf{C}^{\mathrm{geo}}_i$, with $\varepsilon_{\text{min}}=10^{-6}$. These matrices and the base node descriptors are computed once and cached; gradient-dependent quantities and augmented node features are recomputed at each refresh. The superscript distinguishes these geometry-weighted matrices from the unweighted shortest-path matrices used by static EG in~\ref{app:a_static_eg}.

DEG evaluates \paperref{Eq.}{eq:fgw} with the augmented features in \paperref{Eq.}{eq:aug_features} and cached matrices $\mathbf{C}^{\mathrm{geo}}_i$ and $\mathbf{C}^{\mathrm{geo}}_j$, then combines the result with gradient dissimilarity through \paperref{Eq.}{eq:blend}. The FGW structural coefficient is $\alpha=0.5$, distinct from the gradient-mixing coefficient $\alpha_g$ in \paperref{Eq.}{eq:blend}. The solver, Armijo line search, and Ward-linkage qualification are the same as in~\ref{app:a_static_eg}. Ward agglomeration~\citep{ward1963hierarchical} is applied to $D$ without a group-size constraint.

\subsubsection{Soft Routing and Weighted Actor Losses}

For the main FGW variant, \paperref{Eq.}{eq:soft_weights} defines the affinity and normalized robot-to-group weights. The FGW self-affinity is explicitly set to $1$ as a numerical safeguard. With weights fixed during the actor update, the group gradient in \paperref{Eq.}{eq:soft_grad} is the gradient of the normalized weighted actor loss:
\begin{align}
\mathcal{L}^{(m)}(\theta) &= \sum_{i=1}^{N}\bar{w}_{im}\mathcal{L}^{\pi}_i(\theta), \label{eq:group_actor_loss}\\
\bar{w}_{im} &= \frac{w_{im}}{\sum_{j=1}^{N}w_{jm}+\epsilon_g}.
\end{align}
where $\epsilon_g=10^{-8}$.

\subsubsection{Initialization and Refresh Schedule}

Initial groups are formed once from the precomputed pairwise embodiment distances used by static EG (Ward agglomeration; \hyperref[sec:eg_training]{\S\ref*{sec:eg_training}}), and the initial weights are uniform, $w_{im}=1/M$. With these weights, every group gradient is the same and equals the mean of the per-robot gradients up to the $10^{-8}$ denominator stabilizer; therefore the group aggregates are identical until the first successful refresh.

Refresh attempts satisfy $t\geq W_0$ and $t\bmod K=0$; the settings in \paperref{Table}{tab:hyperparameters} give a first attempt at step $5000$. The implementation recomputes per-robot gradients at the post-update parameters using the current iteration's minibatches, then recomputes groups and soft weights according to the selected variant; successful changes take effect in the next training iteration. A refresh is skipped if the largest per-robot gradient norm is below $\delta_g$, leaving the previous groups and weights in place.

At each scheduled monitor/refresh step, the training loop first recomputes per-robot gradients for diagnostics; the regrouping routine then computes them again for the refresh using the same minibatches and updated model parameters.

\subsubsection{Optimization and Training Algorithm}

The value and twin action-value networks retain the IQL losses averaged over all embodiments (\paperref{Section}{sec:backbone}); grouping changes only the actor updates. In each iteration, the actor gradients used for the $M$ sequential group steps are computed once at the same actor parameters and are not recomputed between those steps.

Each weighted group gradient is clipped to global norm $5.0$ and passed to the shared Adam optimizer~\citep{kingma2015adam}. Its cosine learning-rate schedule advances once per group update over the full $T\times M$ update horizon, where $T=100{,}000$ is the number of training iterations. In \paperref{Algorithm}{alg:eg}, $\text{PreDist}_{\mathrm{FGW}}$ denotes the precomputed, normalized morphology-dissimilarity matrix used to initialize static-EG groups.

\begin{algorithm}[H]
\caption{Dynamic Soft Embodiment Grouping (DEG, FGW variant)}
\label{alg:eg}
\begin{algorithmic}[1]
\small\singlespacing
\Require Robots $\mathcal{T}$ with $N=|\mathcal{T}|$, datasets $\{\mathcal{D}_i\}$, iterations $T$, groups $M$, warmup $W_0$, refresh interval $K$
\Require FGW coefficient $\alpha=0.5$, gradient mixing $\alpha_g=0.5$, feature weights $b_1=b_2=1$
\Require Temperature $\kappa=0.1$, gradient threshold $\delta_g=10^{-6}$, target-update rate $\tau_{\mathrm{targ}}=0.005$
\Ensure Trained actor $\theta$, value network $V_\psi$, twin Q-networks $Q_{\phi_1},Q_{\phi_2}$ and target networks $Q_{\bar\phi_1},Q_{\bar\phi_2}$
\State $\{G_m\}_{m=1}^M \gets \text{Ward}(\text{PreDist}_{\mathrm{FGW}})$;\quad $w_{im} \gets 1/M$ \Comment{static-EG groups (\hyperref[sec:eg_training]{\S\ref*{sec:eg_training}}); uniform weights}
\For{$t = 0$ \textbf{to} $T-1$}
    \State Sample one minibatch per robot; update $V_\psi$ using the mean value loss over all batches
    \State $\bm{g}_i \gets \nabla_\theta \mathcal{L}^\pi_i(\theta)$ for all $i$ \Comment{once, at the same actor parameters}
    \For{$m \in \text{perm}(\{1,\dots,M\})$} \Comment{same permutation every iteration}
        \State $\bm{g}^{(m)} \gets \big(\sum_{i=1}^{N} w_{im}\,\bm{g}_i\big) \big/ \big(\sum_{i=1}^{N} w_{im} + 10^{-8}\big)$ \Comment{\paperref{Eq.}{eq:soft_grad}}
        \State $\theta \gets \mathrm{AdamStep}\big(\theta,\ \mathrm{ClipByGlobalNorm}(\bm{g}^{(m)},5.0)\big)$ \Comment{shared optimizer state; cosine schedule over $TM$ steps}
    \EndFor
    \State Update $Q_{\phi_1},Q_{\phi_2}$ using the mean twin-Q loss over all batches
    \State $\bar\phi_1 \gets (1-\tau_{\mathrm{targ}})\bar\phi_1+\tau_{\mathrm{targ}}\phi_1$
    \State $\bar\phi_2 \gets (1-\tau_{\mathrm{targ}})\bar\phi_2+\tau_{\mathrm{targ}}\phi_2$ \Comment{target Q Polyak updates}
    \If{$t \geq W_0$ \textbf{and} $t \bmod K = 0$}
        \State Recompute $\bm{g}^{\mathrm{refresh}}_i=\nabla_\theta\mathcal{L}^{\pi}_i(\theta)$ for all $i$ at post-update parameters
        \If{$\max_i\|\bm{g}^{\mathrm{refresh}}_i\|_2 \geq \delta_g$}
            \State Update groups $\{G_m\}$ and weights $\mathbf{W}$ from refreshed gradients and cached structures \Comment{active from $t{+}1$}
        \EndIf
    \EndIf
\EndFor
\end{algorithmic}
\end{algorithm}

\subsection{Comparison Configurations}
\label{app:a_variants}

These configurations vary grouping, gradient processing, or architecture. The Uniform controls use one actor step per iteration; the DEG ablations are defined in~\ref{app:a_ablations}.

\subsubsection{Direct Gradient-Only Regrouping}

Morphology information is discarded. The cosine-similarity matrix is smoothed across refreshes by an exponential moving average,
\begin{equation}
S^{\mathrm{EMA}}_{r} = \lambda S^{\mathrm{EMA}}_{r-1} + (1-\lambda)S_r, \qquad \lambda=0.9,
\end{equation}
where $r$ indexes refreshes and the first smoothed matrix is the first observed similarity matrix. The Direct distance is $D_{ij}=1-S^{\mathrm{EMA}}_{ij}$. The implementation uses a greedy balanced $k$-medoids variant~\citep{kaufman1990finding} with $k=M$: the first medoid is $\arg\min_j D_{0j}$ (normally robot 0 because $D_{00}=0$), additional medoids are selected farthest-first, and nearest-medoid assignments are capacity-constrained before medoids are updated. Assignment and medoid updates run for at most 20 iterations. For the evaluated $N=16$, $M=4$ setting this gives four equal-sized groups. For soft routing, Direct uses the affinity $\exp(S^{\mathrm{EMA}}_{ij}/\kappa)$ with the group-mass summation and row normalization of \paperref{Eq.}{eq:soft_weights}. Its self-affinity is explicitly set to $\exp(1/\kappa)$, matching $S^{\mathrm{EMA}}_{ii}=1$, as a numerical safeguard.

\subsubsection{Uniform Controls}

The Uniform control averages the actor gradients of all $N$ robots and applies one actor optimizer step per training iteration, without group-specific actor updates. Group-Uniform instead samples one of the fixed static-EG groups uniformly at each iteration and applies one actor step using the mean gradient of that group's robots. Both controls sample minibatches for all robots and retain the value and twin-Q updates averaged over all robots. Gradient projection and dynamic regrouping are disabled in these reported controls. Their actor cosine schedules cover $T$ optimizer steps, compared with $TM$ for sequential grouped updates; they therefore differ from EG and DEG in the number of actor optimizer steps per iteration.

\subsubsection{Geodesic Morphology Distances (GeoEG)}

GeoEG uses geometry-weighted shortest-path distances in the precomputed morphology matrix. GeoEG Static forms its fixed groups from this matrix. The dynamic configurations labeled GeoEG Initialization use the matrix to form their starting groups; subsequent refreshes follow the selected FGW or Direct grouping procedure.

\subsubsection{Gradient Projection: PCGrad and PCGrad-Norm}

PCGrad~\citep{yu2020pcgrad} projects a task gradient sequentially against other conflicting task gradients in a randomized order. The implementation here instead uses a symmetric all-pairs projection: for every conflicting pair, it accumulates both projections computed from the original pair of gradients, then subtracts the accumulated projections before averaging. With $\epsilon_{\mathrm{pc}}=10^{-8}$ for numerical stability, the pairwise modifications are
\begin{equation}
\text{if } \bm{g}_i^\top \bm{g}_j < 0, \quad
\bm{m}_i \mathrel{+}= \frac{\bm{g}_i^\top \bm{g}_j}{\|\bm{g}_j\|_2^2+\epsilon_{\mathrm{pc}}}\bm{g}_j, \quad
\bm{m}_j \mathrel{+}= \frac{\bm{g}_i^\top \bm{g}_j}{\|\bm{g}_i\|_2^2+\epsilon_{\mathrm{pc}}}\bm{g}_i,
\label{eq:pcgrad}
\end{equation}
where $\bm{m}_i$ and $\bm{m}_j$ are initialized to zero, and the projected gradient is $\bm{g}'_i=\bm{g}_i-\bm{m}_i$. Here $\|\cdot\|_2$ denotes the Euclidean norm of the flattened gradient. This implementation can be sensitive to unequal gradient magnitudes.

In PCGrad-Norm, each flattened robot gradient is normalized by $r_i=\sqrt{\|\bm{g}_i\|_2^2+10^{-8}}$, giving $\bm{u}_i=\bm{g}_i/r_i$. For each conflicting pair $i<j$ with $\bm{u}_i^\top\bm{u}_j<0$, the implementation accumulates both projections using the original normalized vectors; after all pairs, it rescales each projected vector by its original norm:
\begin{equation}
\tilde{\bm{u}}_i = \bm{u}_i - \sum_{j\ne i} \mathds{1}(\bm{u}_i^\top\bm{u}_j<0)(\bm{u}_i^\top\bm{u}_j)\bm{u}_j,
\qquad \bm{g}'_i = r_i\tilde{\bm{u}}_i.
\end{equation}
Inter-group projection operates on the $M$ group-level aggregates before sequential stepping, after any within-group projection. The Inter-PCGrad configuration uses the unnormalized all-pairs rule in \paperref{Eq.}{eq:pcgrad}; the configuration labelled PCGrad Norm + Inter-PCGrad uses the normalized rule at both levels.

\subsubsection{Per-Robot Output Heads}

\textbf{Decoupled projection heads} replace the shared actor or critic output projection with robot-specific output parameters (one head per robot):
\begin{align}
\mu_{\theta,\tau}(s) &= W^{\mu}_{\tau}z_{\mu}(s)+b^{\mu}_{\tau}, \qquad
\log\sigma_{\theta,\tau}(s) = W^{\sigma}_{\tau}z_{\sigma}(s)+b^{\sigma}_{\tau}, \label{eq:per_robot_actor} \\
V_{\psi}(s,\tau) &= W^{V}_{\tau}z_{V}(s)+b^{V}_{\tau}, \nonumber \\
Q_{\phi_r}(s,a,\tau) &= W^{Q_r}_{\tau}z_{Q_r}(s,a)+b^{Q_r}_{\tau}, \quad r\in\{1,2\}. \label{eq:per_robot_critic}
\end{align}
The matrices $W$ and biases $b$ are learned robot-specific output parameters. The vectors $z_\mu,z_\sigma,z_V,z_{Q_r}$ are features produced by the shared parts of the actor, value, and Q networks; the actor projections are applied to each joint's feature vector. The actor is a diagonal Gaussian with mean $\mu_{\theta,\tau}$ and standard deviation $\sigma_{\theta,\tau}$; the code clips its mean and log-standard-deviation outputs to configured bounds.

The actor mean head uses the combined actor/action latent, while its log-standard-deviation head uses the joint-description latent. Each value or Q head returns a scalar. We test three configurations: actor-only heads, critic-only heads (both value and twin Q outputs), and both (full per-robot heads).

\subsubsection{Mixture-of-Experts Actor}

\begin{sloppypar}
We evaluate an actor mixture-of-experts architecture adapted from MOORE~\citep{hendawy2024moore}.
\end{sloppypar}

The actor replaces its shared MLP block with $E=4$ parallel experts. The experts receive the actor's shared encoded input $z(s)$, rather than the raw observation.

Each robot has a learned routing embedding $\mathbf{e}_\tau\in\mathbb{R}^{E}$, and its softmax gives the expert weights:
\begin{equation}
h(s, \tau) = \sum_{k=1}^{E} \frac{\exp(e_{\tau,k})}{\sum_{j=1}^{E} \exp(e_{\tau,j})} \, \text{Expert}_k(z(s)).
\label{eq:moore}
\end{equation}
Here $h(s,\tau)$ is the mixed actor latent, $z(s)$ is the actor's encoded input, and each $\text{Expert}_k$ is a two-layer MLP with hidden width 256. The expert networks use orthogonal kernel initialization with gain $\sqrt{2}$. Their dense-layer kernels are re-orthogonalized by QR-based orthogonalization every 10,000 training iterations. The mixture output is fed to the actor's remaining layers; the value and Q networks retain their standard architecture.

This is a MOORE-inspired actor variant, not a direct implementation of MOORE: the original method orthogonalizes expert representations and then combines them with task-specific weights, whereas this code mixes expert outputs using learned robot-specific softmax gates and periodically orthogonalizes expert weight matrices~\citep{hendawy2024moore}.

\subsection{DEG Ablations}
\label{app:a_ablations}

The following ablations separately test continued regrouping and group-specific actor aggregates; they are evaluated apart from the 31-configuration comparison.

\subsubsection{DEG-Freeze}

Training follows DEG up to and including the first successful refresh (\hyperref[sec:exp_grouping]{\S\ref*{sec:exp_grouping}}). After that refresh, all further regrouping events are skipped, retaining the group composition and soft weights for the rest of training. The remaining machinery, including soft routing and the $M$ sequential group steps, is unchanged.

\subsubsection{DEG-Collapse}

The refresh schedule and the soft weights remain active as in DEG. At each iteration, after the $M$ group gradients $\bm{g}^{(m)}$ of \paperref{Eq.}{eq:soft_grad} are computed, each of the $M$ sequential steps applies the same collapsed gradient, the arithmetic mean $\frac{1}{M}\sum_{m=1}^{M}\bm{g}^{(m)}$, instead of its group-specific aggregate, so that no group receives a distinct update. Under uniform weights the $M$ group gradients are identical by construction, so this variant takes full effect only from the first successful refresh onward (\hyperref[sec:exp_grouping]{\S\ref*{sec:exp_grouping}}).

Both variants share the backbone, data, hyperparameters, and evaluation protocol of \hyperref[sec:setup]{\S\ref*{sec:setup}}; their results are reported in \hyperref[sec:results]{\S\ref*{sec:results}}. The implementation switches are \texttt{freeze\_after\_}\allowbreak{}\texttt{first\_regroup} and \texttt{collapse\_}\allowbreak{}\texttt{group\_grads}, respectively.

\clearpage
\section{Soft-Weight Entropy Calculation}
\label{app:entropy}

To measure group-mass concentration in \paperref{Figure}{fig:entropy}, we normalize columns of $\mathbf{W}\in\mathbb{R}^{N\times M}$, the contributions of $N$ robots to $M$ group updates:
\begin{equation}
p_{ig}=\frac{w_{ig}}{\sum_{j=1}^{N}w_{jg}},
\label{eq:entropy_mass}
\end{equation}
where $w_{ig}$ is robot $i$'s contribution to group $g$. The normalized Shannon entropy~\citep{shannon1948i} is then
\begin{equation}
H_g=-\frac{\sum_{i=1}^{N}p_{ig}\log p_{ig}}{\log N}\in[0,1].
\label{eq:entropy_normalized}
\end{equation}
Here $\log$ is the natural logarithm and $0\log0=0$. Entropy approaches zero as mass concentrates on one robot.

The main configuration has $N=16$ and $M=4$. Uniform mass over four robots gives the reference value $\log4/\log16=0.50$, whereas uniform mass over all 16 robots gives $H_g=1$.

For numerical stability, the logger adds $10^{-8}$ to the column sum in \paperref{Eq.}{eq:entropy_mass} and inside $\log(p_{ig}+10^{-8})$ in \paperref{Eq.}{eq:entropy_normalized}. Entropies are recorded to two decimal places.

Across groups, seeds, and refreshes, logged DEG entropies span $0.02$--$0.80$ under unbalanced FGW+Ward clustering, reflecting differentiated, changing groups. The balanced $k$-medoids soft configuration stays within $0.50$--$0.56$, close to the four-robot reference.

\clearpage
\section{Full Comparison Results}
\label{app:full_results}

\paperref{Table}{tab:appendix_full_results} extends \paperref{Table}{tab:main_results} to all 31 configurations, reporting final mean returns, population standard deviations across five seeds, and differences from static EG on 70\% Suboptimal Forward.

\begin{table}[H]
\centering
\scriptsize
\renewcommand{\baselinestretch}{1}\selectfont  
\setlength{\tabcolsep}{2.5pt}
\renewcommand{\arraystretch}{0.85}
\begin{tabular}{lccr}
\toprule
\textbf{Method} & \textbf{Mean} & \textbf{Std} & \textbf{$\Delta$} \\
\midrule
\multicolumn{4}{l}{\textit{Baselines}} \\
\quad Uniform (no groups) & 41.89 & 2.15 & $-$6.39 \\
\quad Group-Uniform (1 group/step) & 31.62 & 7.98 & $-$16.66 \\
\midrule
\multicolumn{4}{l}{\textit{Static grouping}} \\
\quad EG Static & 48.28 & 5.88 & -- \\
\midrule
\multicolumn{4}{l}{\textit{Dynamic grouping}} \\
\quad EG Dynamic Soft (DEG) & 52.79 & 2.13 & $+$4.51 \\
\quad EG Dynamic Direct + Inter-PCGrad & 50.95 & 2.76 & $+$2.67 \\
\quad EG Dynamic Direct & 50.92 & 3.32 & $+$2.64 \\
\quad EG Dynamic (FGW) & 49.81 & 5.57 & $+$1.53 \\
\quad EG Dynamic + PCGrad & 49.56 & 2.54 & $+$1.28 \\
\quad EG Dynamic + PCGrad Norm & 48.52 & 2.96 & $+$0.24 \\
\quad EG Dynamic Direct Soft & 44.79 & 4.81 & $-$3.49 \\
\midrule
\multicolumn{4}{l}{\textit{EG with gradient projection}} \\
\quad EG Dynamic Direct + PCGrad Norm & 50.77 & 2.23 & $+2.49$ \\
\quad EG FGW Soft + PCGrad & 50.35 & 3.17 & $+$2.07 \\
\quad EG FGW Soft + PCGrad Norm & 48.91 & 4.32 & $+$0.63 \\
\quad EG Dynamic Direct + PCGrad & 48.90 & 5.19 & $+0.62$ \\
\quad EG Static + PCGrad & 48.71 & 2.96 & $+$0.43 \\
\quad EG Static + PCGrad Norm & 47.91 & 3.07 & $-$0.37 \\
\quad EG Dynamic Direct + PCGrad Norm + Inter-PCGrad & 47.25 & 3.08 & $-1.03$ \\
\quad EG Dynamic Direct Soft + PCGrad & 44.82 & 6.16 & $-$3.46 \\
\midrule
\multicolumn{4}{l}{\textit{Architecture variants}} \\
\quad EG Static + Per-Robot Heads & 50.05 & 1.80 & $+$1.77 \\
\quad EG Static + Per-Robot Actor Heads & 49.46 & 1.75 & $+$1.18 \\
\quad EG Static + Per-Robot Critic Heads & 48.47 & 2.82 & $+$0.19 \\
\quad EG Static + MoE & 30.66 & 2.94 & $-$17.62 \\
\midrule
\multicolumn{4}{l}{\textit{GeoEG (Static)}} \\
\quad GeoEG Static + Per-Robot Actor Heads & 46.65 & 2.10 & $-$1.63 \\
\quad GeoEG Static + PCGrad & 46.30 & 3.29 & $-$1.98 \\
\quad GeoEG Static & 45.92 & 5.14 & $-$2.36 \\
\quad GeoEG Static + PCGrad Norm & 44.54 & 2.79 & $-$3.74 \\
\quad GeoEG Static + Per-Robot Critic Heads & 44.32 & 4.28 & $-$3.96 \\
\quad GeoEG Static + Per-Robot Heads & 42.04 & 1.16 & $-$6.24 \\
\quad GeoEG Static + MoE & 29.71 & 3.00 & $-$18.57 \\
\midrule
\multicolumn{4}{l}{\textit{Dynamic EG with GeoEG Initialization}} \\
\quad EG Dynamic + GeoEG Initialization & 43.20 & 3.09 & $-5.08$ \\
\quad EG Dynamic Direct + GeoEG Initialization & 42.97 & 3.86 & $-5.31$ \\
\bottomrule
\end{tabular}
\caption{Full reported comparison of 31 configurations on 70\% Suboptimal Forward @100K. Methods grouped by category and sorted by mean return. Std across 5 seeds (population, ddof=0). $\Delta$ is the difference from EG Static.}
\label{tab:appendix_full_results}
\end{table}

\clearpage
\section{Additional Learning Curves}
\label{app:additional_curves}

\begin{figure}[htbp]
\centering
\includegraphics[width=0.98\linewidth]{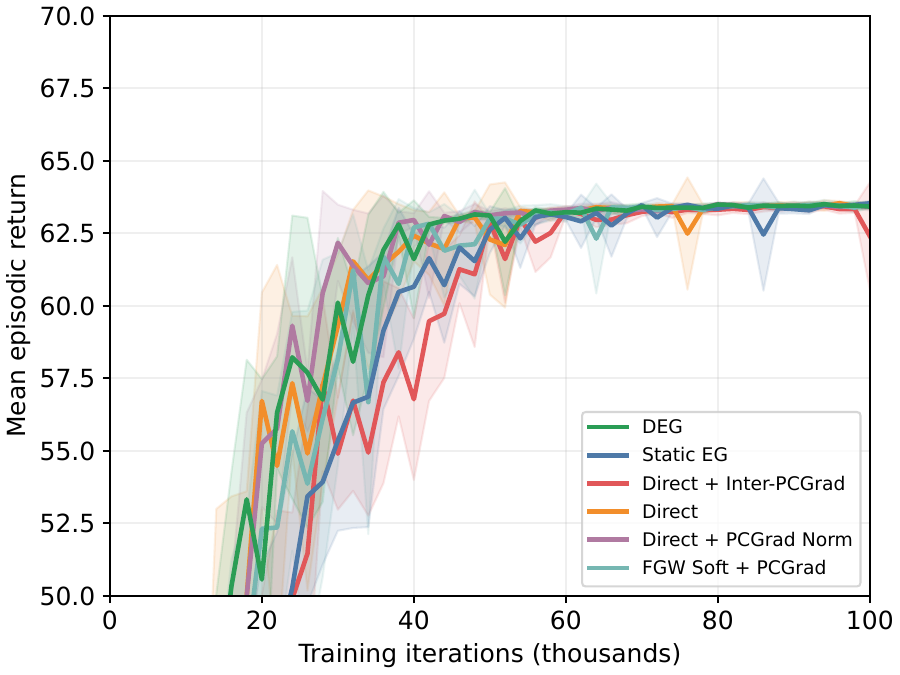}
\caption{Learning curves on Expert Forward, recorded every 2K iterations. The selected methods have similar final mean returns; Inter-PCGrad reaches 62.40.}
\label{fig:app_expert_curves}
\end{figure}

\begin{figure}[htbp]
\centering
\includegraphics[width=0.98\linewidth]{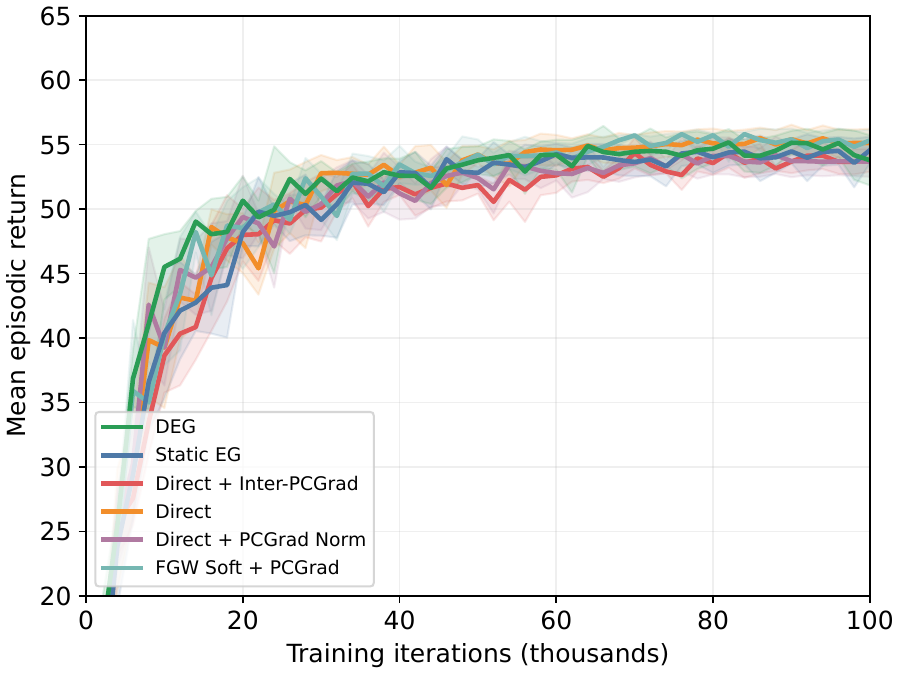}
\caption{Learning curves on Replay Forward, recorded every 2K iterations. The relative ordering changes during training; EG FGW Soft+PCGrad has the highest final mean among the methods shown.}
\label{fig:app_replay_curves}
\end{figure}

\paperref{Figures}{fig:app_sub70_curves}, \ref{fig:app_expert_curves}, and~\ref{fig:app_replay_curves} show six selected methods across the three dataset variants. Each method is plotted only at checkpoints recorded for all five of its seeds. Lines connect these observations; shaded regions show $\pm1$ population standard deviation across seeds.

\begin{figure}[htbp]
\centering
\includegraphics[width=0.98\linewidth]{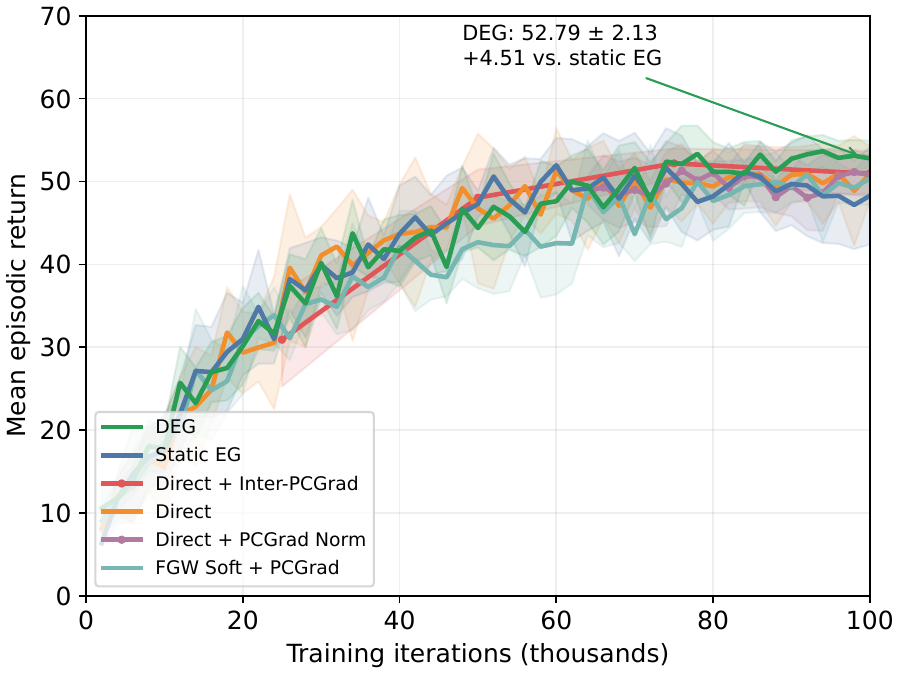}
\caption{Learning curves on 70\% Suboptimal Forward. DEG has the highest final mean among the methods shown. Inter-PCGrad has four shared checkpoints (25K--100K), and PCGrad Norm begins at 64K; the other methods have 2K spacing.}
\label{fig:app_learning_curves}
\label{fig:app_sub70_curves}
\end{figure}

\clearpage
\begin{landscape}
\section{Per-Robot Results on Suboptimal Data}
\label{app:per_robot}

\paperref{Table}{tab:appendix_per_robot} reports mean episodic returns for each of the 16 robots across the 31 reported configurations. Each value averages the final evaluation return over 5 seeds.

\begin{table}[H]
\centering
\setlength{\tabcolsep}{3pt}
\renewcommand{\arraystretch}{0.85}
\caption{Per-robot mean episodic return for the 31 reported configurations on 70\% Suboptimal Forward @100K. Values averaged across 5 seeds.}
\label{tab:appendix_per_robot}
\resizebox{\linewidth}{!}{%
\begin{tabular}{lrrrrrrrrrrrrrrrrr}
\toprule
Method & a1 & go1 & go2 & anymal\_b & anymal\_c & bkv0 & bkvb & badger & bittle & h1 & g1 & talos & op3 & nao\_v5 & cassie & hexapod & Mean \\
\midrule
EG Dynamic Soft (DEG) & 37.7 & 45.1 & 52.4 & 48.2 & 43.6 & 46.8 & 55.0 & 52.7 & 45.2 & 53.8 & 62.0 & 109.7 & 101.5 & 89.1 & 1.3 & 0.4 & 52.8 \\
EG Dynamic + PCGrad & 30.0 & 37.3 & 52.4 & 48.2 & 43.6 & 46.5 & 55.0 & 52.5 & 45.4 & 22.2 & 75.7 & 110.4 & 100.2 & 71.2 & 1.4 & 0.7 & 49.6 \\
EG FGW Soft + PCGrad & 22.2 & 37.4 & 52.4 & 48.1 & 43.8 & 46.6 & 55.0 & 52.6 & 45.2 & 43.2 & 76.6 & 107.9 & 101.1 & 71.9 & 1.4 & 0.1 & 50.3 \\
EG Static + Per-Robot Heads & 22.2 & 45.0 & 52.3 & 48.1 & 43.6 & 46.7 & 55.0 & 52.6 & 44.9 & 54.4 & 32.3 & 110.3 & 102.1 & 89.7 & 1.3 & 0.3 & 50.0 \\
EG Dynamic (FGW) & 29.9 & 45.1 & 52.4 & 48.0 & 43.5 & 46.3 & 55.0 & 52.8 & 45.0 & 33.0 & 61.2 & 110.5 & 100.9 & 71.7 & 1.3 & 0.3 & 49.8 \\
EG Static + Per-Robot Actor Heads & 30.0 & 52.8 & 52.3 & 48.2 & 43.8 & 46.1 & 55.0 & 52.6 & 45.3 & 2.3 & 77.5 & 111.3 & 100.7 & 71.5 & 1.4 & 0.5 & 49.5 \\
EG FGW Soft + PCGrad Norm & 22.2 & 22.1 & 52.4 & 48.2 & 43.8 & 46.7 & 55.0 & 52.6 & 45.0 & 43.5 & 49.0 & 110.3 & 100.8 & 89.5 & 1.3 & 0.2 & 48.9 \\
EG Static + PCGrad & 22.2 & 22.1 & 52.4 & 48.1 & 43.6 & 46.8 & 55.0 & 52.5 & 45.3 & 32.9 & 54.6 & 111.1 & 101.5 & 89.3 & 1.3 & 0.6 & 48.7 \\
EG Dynamic + PCGrad Norm & 22.2 & 37.5 & 52.4 & 48.1 & 43.5 & 46.7 & 55.0 & 52.7 & 44.8 & 33.0 & 76.2 & 89.2 & 101.3 & 72.2 & 1.3 & 0.3 & 48.5 \\
EG Static + Per-Robot Critic Heads & 14.4 & 22.1 & 52.4 & 48.3 & 43.5 & 46.5 & 54.9 & 52.7 & 44.9 & 54.3 & 37.7 & 110.7 & 102.0 & 89.4 & 1.4 & 0.2 & 48.5 \\
EG Static & 22.2 & 37.4 & 52.4 & 48.2 & 43.5 & 46.7 & 55.0 & 52.6 & 45.2 & 43.6 & 61.4 & 89.3 & 101.3 & 71.9 & 1.4 & 0.5 & 48.3 \\
EG Static + PCGrad Norm & 22.2 & 22.2 & 52.4 & 48.1 & 43.5 & 46.8 & 55.0 & 52.5 & 45.0 & 11.8 & 64.6 & 111.2 & 100.1 & 89.0 & 1.4 & 0.5 & 47.9 \\
EG Dynamic + GeoEG Initialization & 22.2 & 37.4 & 52.4 & 48.0 & 43.5 & 46.8 & 55.0 & 52.6 & 45.3 & 31.9 & 78.0 & 21.6 & 101.1 & 53.9 & 1.5 & 0.2 & 43.2 \\
EG Dynamic Direct + GeoEG Initialization & 22.2 & 29.8 & 52.4 & 48.2 & 43.4 & 46.9 & 55.0 & 52.6 & 45.3 & 22.5 & 78.4 & 34.7 & 100.8 & 53.8 & 1.4 & 0.3 & 43.0 \\
EG Dynamic Direct + PCGrad Norm + Inter-PCGrad & 22.2 & 37.5 & 52.4 & 48.1 & 43.6 & 46.3 & 55.0 & 52.7 & 45.0 & 54.2 & 46.6 & 95.1 & 101.3 & 54.0 & 1.4 & 0.6 & 47.2 \\
GeoEG Static + Per-Robot Actor Heads & 14.5 & 52.6 & 52.3 & 48.1 & 43.2 & 46.3 & 55.0 & 52.6 & 45.5 & 32.5 & 78.0 & 33.3 & 101.5 & 89.1 & 1.3 & 0.5 & 46.6 \\
EG Dynamic Direct + PCGrad Norm & 22.1 & 29.8 & 52.4 & 48.2 & 43.6 & 46.8 & 55.0 & 52.6 & 45.0 & 43.4 & 71.0 & 110.9 & 100.7 & 88.9 & 1.3 & 0.7 & 50.8 \\
GeoEG Static + PCGrad & 29.9 & 37.5 & 52.4 & 47.9 & 43.4 & 46.4 & 55.0 & 52.6 & 45.4 & 21.5 & 78.3 & 38.7 & 101.1 & 89.2 & 1.3 & 0.2 & 46.3 \\
EG Dynamic Direct + Inter-PCGrad & 22.2 & 29.7 & 52.3 & 48.1 & 43.6 & 46.7 & 55.0 & 52.6 & 45.0 & 54.4 & 61.9 & 110.8 & 101.6 & 89.6 & 1.3 & 0.4 & 50.9 \\
EG Dynamic Direct & 22.1 & 37.5 & 52.4 & 48.1 & 43.6 & 46.6 & 55.0 & 52.7 & 45.3 & 31.8 & 76.9 & 111.0 & 100.8 & 89.3 & 1.3 & 0.3 & 50.9 \\
GeoEG Static & 22.2 & 37.5 & 52.4 & 48.2 & 43.4 & 46.4 & 55.0 & 52.6 & 45.4 & 42.4 & 78.4 & 36.3 & 101.3 & 71.3 & 1.4 & 0.5 & 45.9 \\
EG Dynamic Direct Soft + PCGrad & 30.0 & 37.3 & 52.4 & 48.1 & 43.6 & 46.9 & 55.0 & 52.7 & 45.4 & 32.5 & 46.4 & 90.4 & 81.2 & 53.8 & 1.3 & 0.2 & 44.8 \\
EG Dynamic Direct Soft & 37.7 & 37.5 & 52.3 & 48.2 & 43.6 & 46.7 & 55.0 & 52.6 & 45.3 & 21.7 & 34.0 & 67.9 & 100.6 & 71.9 & 1.4 & 0.2 & 44.8 \\
GeoEG Static + PCGrad Norm & 30.0 & 37.5 & 52.4 & 48.0 & 43.5 & 46.8 & 55.0 & 52.7 & 45.5 & 22.4 & 77.5 & 27.2 & 101.3 & 71.2 & 1.5 & 0.2 & 44.5 \\
GeoEG Static + Per-Robot Critic Heads & 22.2 & 29.8 & 52.4 & 48.1 & 43.5 & 46.5 & 55.0 & 52.6 & 45.3 & 32.4 & 77.7 & 29.2 & 101.2 & 71.4 & 1.5 & 0.2 & 44.3 \\
EG Dynamic Direct + PCGrad & 22.2 & 37.5 & 52.4 & 48.2 & 43.5 & 46.3 & 55.0 & 52.7 & 45.1 & 54.0 & 61.3 & 89.1 & 101.6 & 72.0 & 1.2 & 0.3 & 48.9 \\
GeoEG Static + Per-Robot Heads & 14.5 & 22.2 & 52.4 & 48.0 & 43.4 & 46.6 & 55.0 & 52.7 & 45.3 & 11.9 & 78.2 & 10.0 & 101.2 & 89.3 & 1.5 & 0.5 & 42.0 \\
Uniform Baseline & 14.4 & 44.9 & 52.1 & 47.4 & 43.2 & 46.2 & 54.9 & 37.9 & 45.0 & 26.8 & 45.8 & 22.2 & 99.3 & 88.4 & 1.6 & 0.1 & 41.9 \\
Group-Uniform Baseline & 22.1 & 22.0 & 51.9 & 47.2 & 41.9 & 43.9 & 54.3 & 38.0 & 44.7 & 11.2 & 20.4 & 18.1 & 53.8 & 34.7 & 1.4 & 0.2 & 31.6 \\
EG Static + MoE & 45.4 & 52.4 & 52.0 & 39.1 & 1.3 & 43.6 & 54.7 & 52.7 & 43.5 & 1.0 & 3.9 & 5.5 & 75.6 & 18.1 & 1.6 & 0.1 & 30.7 \\
GeoEG Static + MoE & 37.3 & 52.6 & 52.0 & 46.3 & 9.6 & 44.5 & 54.7 & 52.7 & 43.8 & 0.9 & 16.7 & 3.3 & 41.0 & 17.9 & 1.7 & 0.1 & 29.7 \\
\bottomrule
\end{tabular}
}
\end{table}

\end{landscape}
\clearpage
\section{Per-Robot Results on Expert and Replay Data}
\label{app:per_robot_datasets}

\paperref{Tables}{tab:appendix_per_robot_expert} and~\ref{tab:appendix_per_robot_replay} report the per-robot mean episodic return for the seven methods evaluated on expert demonstrations and replay forward data (\paperref{Section}{sec:results}, \paperref{Table}{tab:data_quality}). Each value is the average of the final evaluation return over 5 seeds.

\begin{figure}[htbp]
\centering
\includegraphics[width=0.98\linewidth]{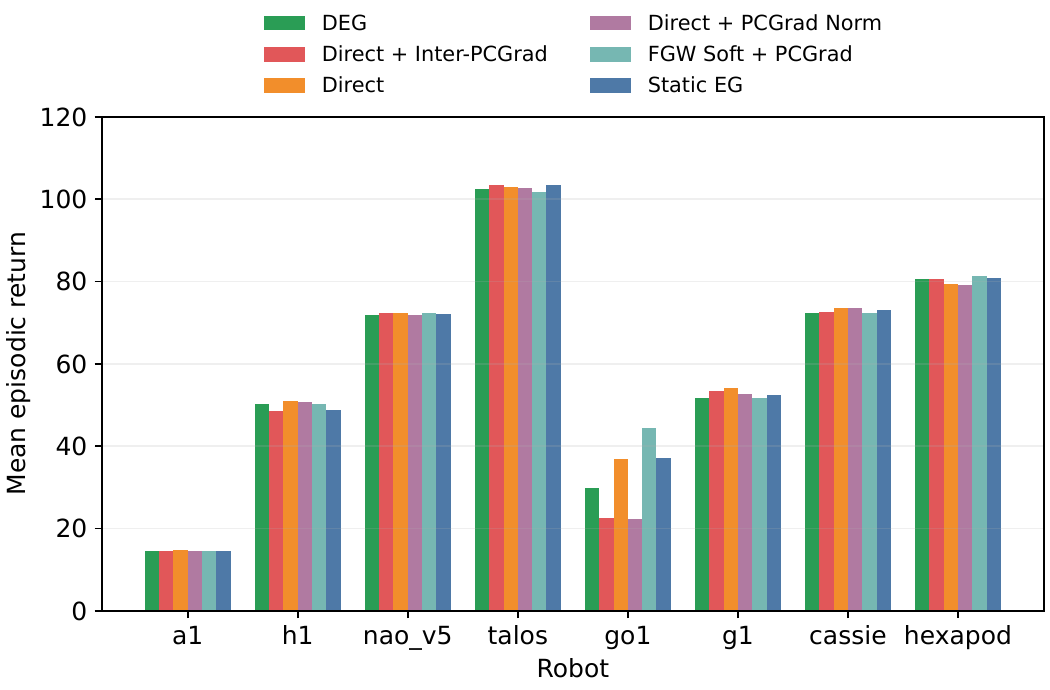}
\caption{Mean episodic returns for eight selected robots and six methods on Replay Forward, averaged over five seeds at 100K iterations. Cassie and hexapod have higher returns on this dataset than on the suboptimal dataset.}
\label{fig:per_robot_replay}
\end{figure}

\paperref{Figures}{fig:per_robot_expert} and~\ref{fig:per_robot_replay} visualize eight selected robots for six methods on expert and replay data, respectively.

\begin{figure}[htbp]
\centering
\includegraphics[width=0.98\linewidth]{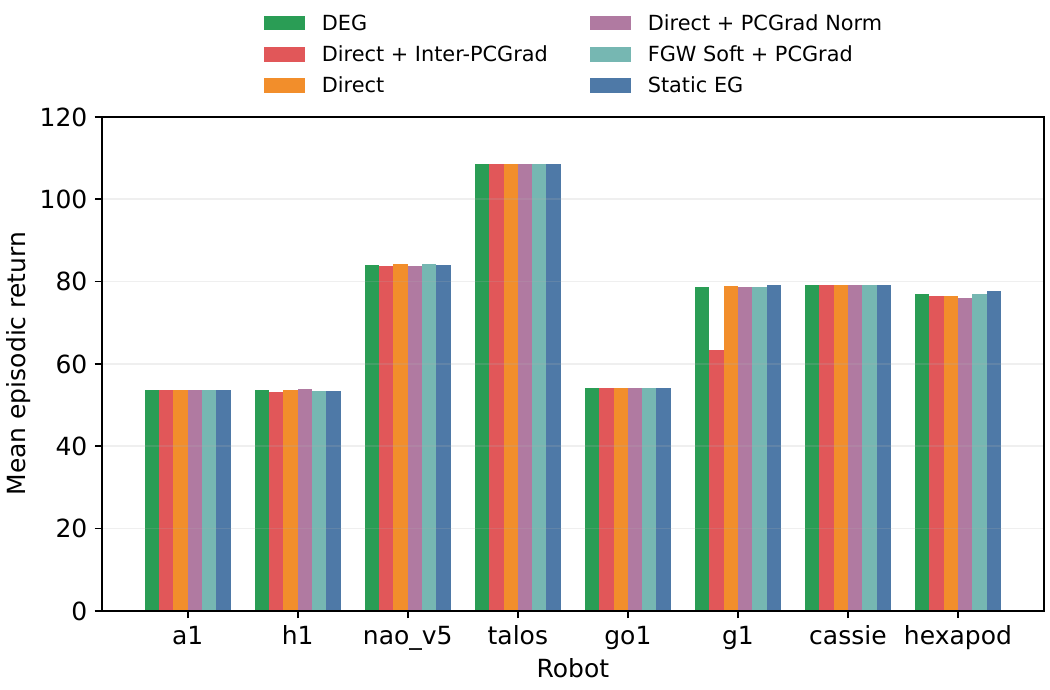}
\caption{Mean episodic returns for eight selected robots and six methods on Expert Forward. Values average five seeds at 100K iterations.}
\label{fig:per_robot_expert}
\end{figure}

\clearpage
\begin{landscape}
\begin{table}[H]
\centering
\setlength{\tabcolsep}{3pt}
\renewcommand{\arraystretch}{0.95}
\caption{Per-robot mean episodic return for all methods on Expert Forward @100K. Values averaged across 5 seeds.}
\label{tab:appendix_per_robot_expert}
\resizebox{\linewidth}{!}{%
\begin{tabular}{lrrrrrrrrrrrrrrrrr}
\toprule
Method & a1 & go1 & go2 & anymal\_b & anymal\_c & bkv0 & bkvb & badger & bittle & h1 & g1 & talos & op3 & nao\_v5 & cassie & hexapod & Mean \\
\midrule
EG Dynamic Soft (DEG) & 53.7 & 54.0 & 53.5 & 49.8 & 44.4 & 46.7 & 53.3 & 52.9 & 37.3 & 53.6 & 78.7 & 108.5 & 88.4 & 84.0 & 79.1 & 76.9 & 63.4 \\
EG Dynamic Soft ($M{=}7$) & 53.7 & 54.0 & 53.5 & 49.8 & 44.4 & 46.7 & 53.4 & 52.9 & 37.8 & 53.1 & 78.8 & 108.5 & 88.6 & 83.6 & 79.1 & 75.3 & 63.3 \\
EG FGW Soft + PCGrad & 53.7 & 54.0 & 53.5 & 49.8 & 44.4 & 46.6 & 53.3 & 52.9 & 38.0 & 53.5 & 78.7 & 108.4 & 88.9 & 84.1 & 79.1 & 76.9 & 63.5 \\
EG Dynamic Direct + PCGrad Norm & 53.7 & 54.0 & 53.5 & 49.8 & 44.4 & 46.6 & 53.4 & 52.9 & 37.3 & 53.8 & 78.7 & 108.4 & 89.1 & 83.7 & 79.1 & 76.0 & 63.4 \\
EG Dynamic Direct + Inter-PCGrad & 53.7 & 54.0 & 53.5 & 49.8 & 44.4 & 46.5 & 53.4 & 52.9 & 37.8 & 53.1 & 63.3 & 108.4 & 88.4 & 83.8 & 79.1 & 76.3 & 62.4 \\
EG Dynamic Direct & 53.7 & 54.0 & 53.5 & 49.8 & 44.5 & 46.9 & 53.4 & 52.9 & 37.3 & 53.7 & 78.9 & 108.5 & 88.9 & 84.1 & 79.1 & 76.6 & 63.5 \\
EG Static & 53.7 & 54.0 & 53.5 & 49.8 & 44.4 & 46.5 & 53.5 & 52.9 & 38.0 & 53.5 & 79.1 & 108.4 & 88.3 & 84.1 & 79.1 & 77.6 & 63.5 \\
\bottomrule
\end{tabular}
}
\end{table}

\begin{table}[H]
\centering
\setlength{\tabcolsep}{3pt}
\renewcommand{\arraystretch}{0.95}
\caption{Per-robot mean episodic return for all methods on Replay Forward @100K. Values averaged across 5 seeds.}
\label{tab:appendix_per_robot_replay}
\resizebox{\linewidth}{!}{%
\begin{tabular}{lrrrrrrrrrrrrrrrrr}
\toprule
Method & a1 & go1 & go2 & anymal\_b & anymal\_c & bkv0 & bkvb & badger & bittle & h1 & g1 & talos & op3 & nao\_v5 & cassie & hexapod & Mean \\
\midrule
EG Dynamic Soft (DEG) & 14.6 & 29.7 & 33.8 & 45.2 & 38.9 & 38.4 & 53.7 & 43.9 & 42.3 & 50.1 & 51.6 & 102.4 & 91.2 & 71.8 & 72.4 & 80.6 & 53.8 \\
EG Dynamic Soft ($M{=}7$) & 14.6 & 44.2 & 42.3 & 45.1 & 38.8 & 38.6 & 53.5 & 45.1 & 42.1 & 49.6 & 50.4 & 102.2 & 90.5 & 71.3 & 72.7 & 81.0 & 55.1 \\
EG FGW Soft + PCGrad & 14.6 & 44.5 & 42.0 & 45.1 & 39.2 & 38.3 & 53.8 & 45.2 & 42.1 & 50.1 & 51.7 & 101.7 & 91.2 & 72.3 & 72.2 & 81.4 & 55.3 \\
EG Dynamic Direct + PCGrad Norm & 14.6 & 22.4 & 42.3 & 45.4 & 39.0 & 38.5 & 53.8 & 38.3 & 42.5 & 50.8 & 52.5 & 102.8 & 91.1 & 71.9 & 73.5 & 79.2 & 53.7 \\
EG Dynamic Direct + Inter-PCGrad & 14.6 & 22.4 & 42.4 & 45.3 & 39.3 & 38.5 & 53.7 & 41.1 & 42.1 & 48.6 & 53.4 & 103.4 & 90.5 & 72.4 & 72.5 & 80.6 & 53.8 \\
EG Dynamic Direct & 14.7 & 37.0 & 43.5 & 45.3 & 39.2 & 38.7 & 53.6 & 42.7 & 42.0 & 50.9 & 54.1 & 102.8 & 92.0 & 72.4 & 73.5 & 79.4 & 55.1 \\
EG Static & 14.6 & 37.0 & 42.1 & 45.2 & 39.3 & 38.6 & 53.7 & 39.9 & 42.3 & 48.7 & 52.4 & 103.4 & 90.1 & 72.0 & 73.0 & 80.8 & 54.6 \\
\bottomrule
\end{tabular}
}
\end{table}

\end{landscape}
\clearpage
\section{Rescued Robots and Group Evolution}
\label{app:rescued}

\begin{figure}[htbp]
\centering
\includegraphics[width=0.98\linewidth]{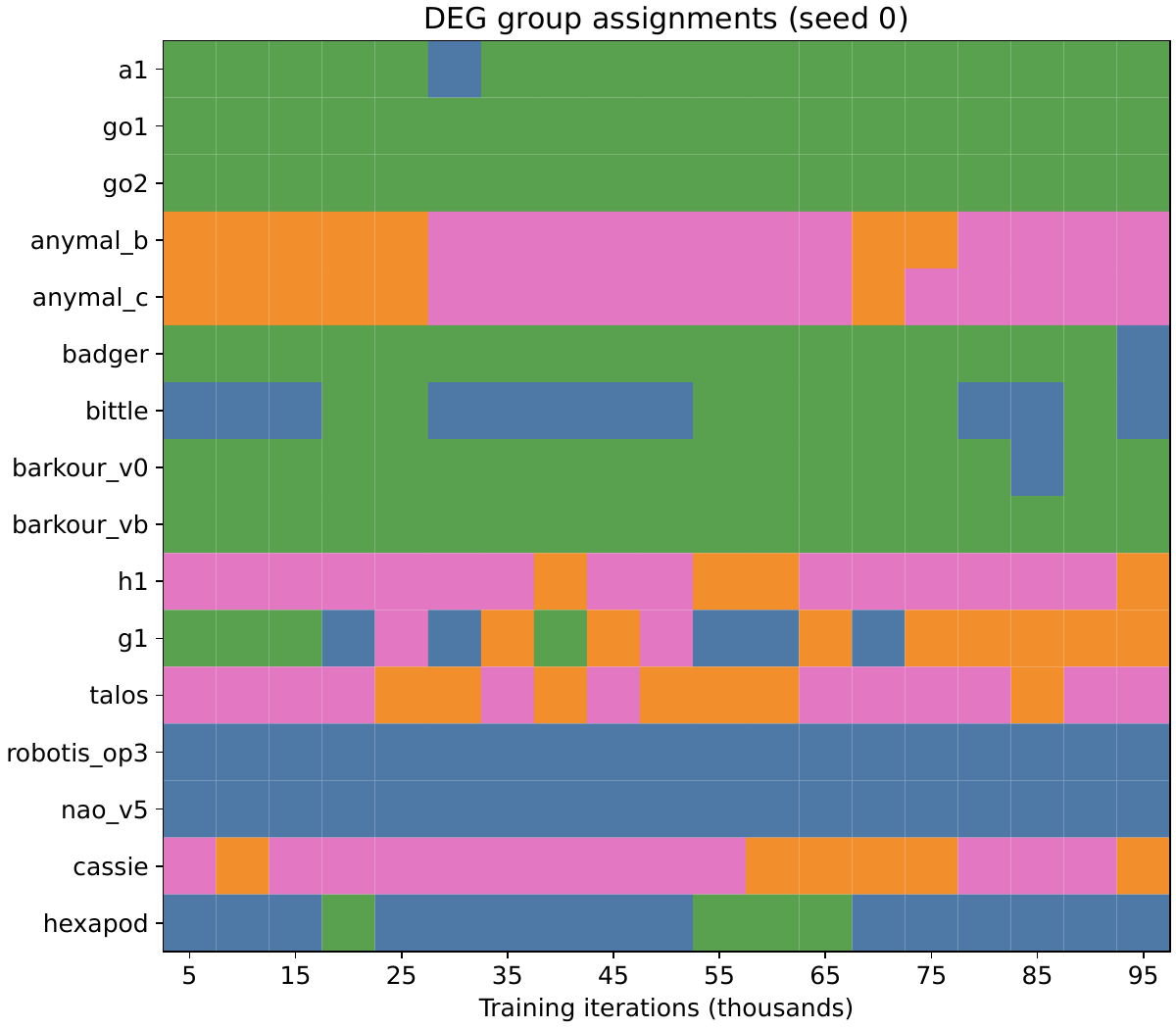}
\caption{DEG group assignments on 70\% Suboptimal Forward (seed~0). Group labels are matched between consecutive refreshes to maximize membership overlap, so color changes indicate changes in robot membership rather than arbitrary cluster relabeling.}
\label{fig:group_evolution}
\end{figure}

\paperref{Figures}{fig:rescued_robots} and~\ref{fig:group_evolution} show the rescue counts across five seeds and the seed-0 group assignments over training, respectively.

\begin{figure}[htbp]
\centering
\includegraphics[width=0.98\linewidth]{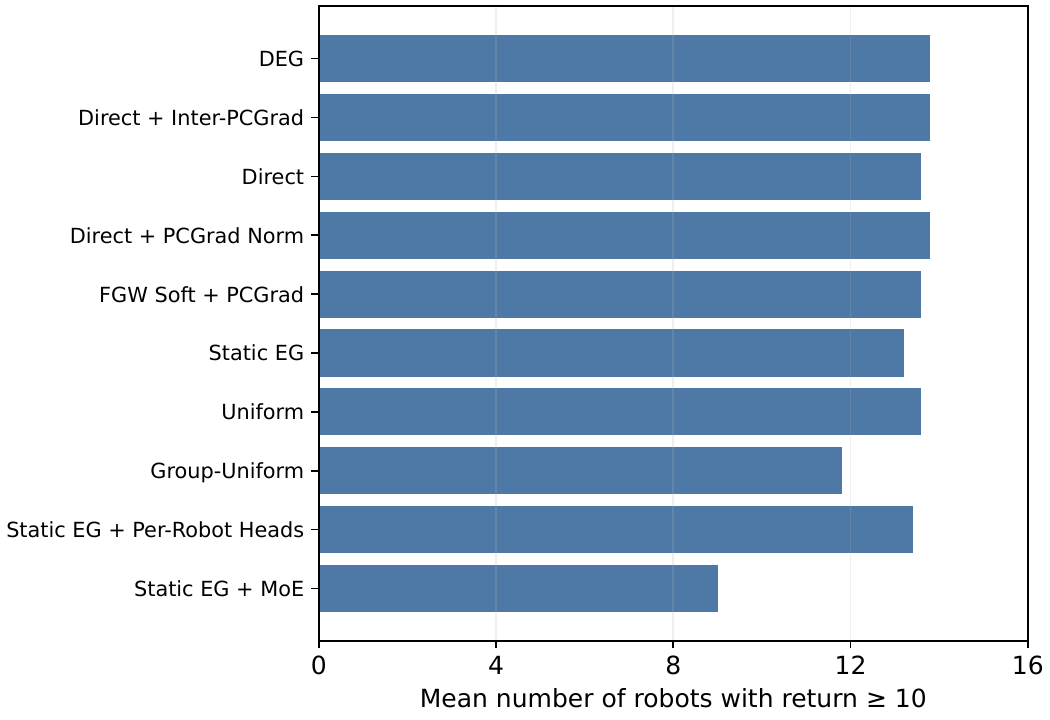}
\caption{Mean number of robots rescued (return $\geq 10$) by each method on 70\% Suboptimal data across 5 seeds. DEG averages 13.8 rescued robots, while the adapted MoE variant averages 9.0.}
\label{fig:rescued_robots}
\end{figure}

\clearpage
\section{Training and Evaluation Details}
\label{app:training}

\paperref{Table}{tab:appendix_protocol} gives the default network architecture and common evaluation protocol for the URMA-based models~\citep{bohlinger2024urma} and EG update framework~\citep{abe2024eg}. \paperref{Table}{tab:hyperparameters} collects the training, grouping, variant, and numerical settings used with the definitions in~\ref{app:methods}. Computational cost is analyzed in~\ref{app:cost}.

\begin{table}[H]
\centering
\footnotesize\singlespacing
\renewcommand{\arraystretch}{1.15}
\begin{tabular}{@{}>{\raggedright\arraybackslash}p{0.24\textwidth}@{\hspace{0.04\textwidth}}>{\raggedright\arraybackslash}p{0.72\textwidth}@{}}
\toprule
\textbf{Component} & \textbf{Specification} \\
\midrule
Observation encoders & Joint descriptor networks use two 64-unit layers and 4-dimensional state projections; foot descriptor networks use two 32-unit layers and 4-dimensional state projections. Their aggregated latents have dimensions 256 and 128, respectively. \\\addlinespace[5pt]
Actor and value networks & URMA joint/foot attention encoders followed by three hidden layers of widths 512, 256, and 128; LayerNorm and ELU are used at the implementation-specified layers. \\\addlinespace[5pt]
Actor output features & Two 128-unit joint-description layers and a 128-unit mean-feature layer precede scalar projections for each valid action coordinate. \\\addlinespace[5pt]
Q critics & Two independently parameterized critics, each with URMA encoders followed by four hidden layers of widths 512, 256, 256, and 256 and a scalar output layer; target copies are maintained. \\\addlinespace[5pt]
Evaluation schedule & 50 scheduled evaluations, every 2K training iterations; 30 episodes per robot at each evaluation; deterministic policy-mean actions. \\\addlinespace[5pt]
Aggregation & Final mean return across 16 robots; population standard deviation across seeds 0--4. A rescued robot has mean return $\geq10$. \\
\bottomrule
\end{tabular}
\caption{Default network architecture and common evaluation protocol. Output-head and MoE variants are defined in~\ref{app:a_variants}; numerical settings are collected in \paperref{Table}{tab:hyperparameters}.}
\label{tab:appendix_protocol}
\end{table}

\clearpage
\begin{table}[H]
\centering
\footnotesize\singlespacing
\renewcommand{\arraystretch}{1.12}
\caption{Hyperparameters and numerical settings for the reported configurations. Shared settings apply unless a comparison or ablation explicitly changes the mechanism. Variant-specific settings are identified below. Adam moment/stability and FGW termination settings use library defaults.}
\label{tab:hyperparameters}
\begin{tabular}{@{}>{\raggedright\arraybackslash}p{0.20\textwidth}@{\hspace{0.03\textwidth}}>{\raggedright\arraybackslash}p{0.47\textwidth}@{\hspace{0.03\textwidth}}>{\raggedright\arraybackslash}p{0.27\textwidth}@{}}
\toprule
\textbf{Parameter} & \textbf{Role} & \textbf{Value or setting} \\
\midrule
\multicolumn{3}{@{}l}{\textbf{Shared optimization and IQL}}\\[4pt]
Learning rates & Initial actor rate; constant value and twin-Q rates & $3\times10^{-4}$ each \\\addlinespace[2pt]
Batch size & Transitions per robot per iteration & $1024$ \\\addlinespace[2pt]
$T$ & Training iterations & $100{,}000$ \\\addlinespace[2pt]
Adam moments & First- and second-moment decay coefficients & $0.9$, $0.999$ \\\addlinespace[2pt]
Adam stability & Denominator epsilon; epsilon inside the square root & $10^{-8}$; $0$ \\\addlinespace[2pt]
Gradient clipping & Maximum global gradient norm & $5.0$ \\\addlinespace[2pt]
Actor schedule & Cosine decay over optimizer steps & $TM$ for sequential groups; $T$ for Uniform controls \\\addlinespace[2pt]
$\omega$ & Value expectile & $0.7$ \\\addlinespace[2pt]
$\beta$ & Exponential advantage-weight coefficient & $3.0$ \\\addlinespace[2pt]
$c_A$ & Maximum dataset-action weight & $100$ \\\addlinespace[2pt]
$\gamma$ & Discount factor & $0.99$ \\\addlinespace[2pt]
$\tau_{\mathrm{targ}}$ & Target-Q Polyak update rate & $0.005$ \\\addlinespace[2pt]
Robot weights & Relative loss and minibatch weights & $1$ for each robot \\\addlinespace[2pt]
State normalization & Additional dataset-state normalization & Disabled \\\addlinespace[5pt]
\multicolumn{3}{@{}l}{\textbf{FGW grouping and soft routing}}\\[4pt]
$M$ & Main group count; cross-dataset alternative & $4$; $7$ \\\addlinespace[2pt]
$\alpha$ & FGW structural coefficient & $0.5$ \\\addlinespace[2pt]
$\alpha_g$ & Graph/gradient distance mixing coefficient & $0.5$ \\\addlinespace[2pt]
$b_1,b_2$ & Scales for mean similarity and its change & $1$, $1$ \\\addlinespace[2pt]
$\kappa$ & Soft-routing temperature & $0.1$ \\\addlinespace[2pt]
$K$ & Refresh interval, in training iterations & $5000$ \\\addlinespace[2pt]

\bottomrule
\end{tabular}
\end{table}

\clearpage
\begin{table}[H]
\ContinuedFloat
\centering
\footnotesize\singlespacing
\renewcommand{\arraystretch}{1.12}
\caption{Hyperparameters and numerical settings (continued).}
\begin{tabular}{@{}>{\raggedright\arraybackslash}p{0.20\textwidth}@{\hspace{0.03\textwidth}}>{\raggedright\arraybackslash}p{0.47\textwidth}@{\hspace{0.03\textwidth}}>{\raggedright\arraybackslash}p{0.27\textwidth}@{}}
\toprule
\textbf{Parameter} & \textbf{Role} & \textbf{Value or setting} \\
\midrule
\multicolumn{3}{@{}l}{\textbf{FGW grouping and soft routing (continued)}}\\[4pt]
$W_0$ & Warmup, in training iterations & $1000$; first attempt at $5000$ \\\addlinespace[1.5pt]
$\delta_g$ & Largest-gradient-norm threshold for refreshing & $10^{-6}$ \\\addlinespace[1.5pt]
FGW solver & Conditional-gradient optimization & Armijo line search \\\addlinespace[1.5pt]
FGW termination & Maximum solver iterations; relative/absolute tolerances & $10{,}000$; $10^{-9}$, $10^{-9}$ (POT defaults) \\\addlinespace[5pt]
\multicolumn{3}{@{}l}{\textbf{Comparison configurations}}\\[4pt]
$\lambda$ & Direct cosine-similarity EMA decay & $0.9$ \\\addlinespace[1.5pt]
Medoid iterations & Maximum balanced $k$-medoids iterations & $20$ \\\addlinespace[1.5pt]
$E$ & MoE actor expert count & $4$ \\\addlinespace[1.5pt]
Expert width & Hidden units in each of two expert layers & $256$ \\\addlinespace[1.5pt]
Kernel refresh & MoE expert-kernel re-orthogonalization interval & $10{,}000$ iterations \\\addlinespace[1.5pt]
Gate initialization & Normal standard deviation for robot-routing embeddings & $0.1$ \\\addlinespace[1.5pt]

\multicolumn{3}{@{}l}{\textbf{Policy bounds and numerical safeguards}}\\[4pt]
Attention temperature & Learned encoder temperature: initial value; lower offset & $1.0$; $0.015$ \\\addlinespace[1.5pt]
Policy scale & Initial standard-deviation bias scale & $1.0$ \\\addlinespace[1.5pt]
Policy bounds & Mean range; standard-deviation range & $[-10,10]$; $[10^{-8},2]$ \\\addlinespace[1.5pt]
$\epsilon_{\mathrm{cos}},\epsilon_{\mathrm{norm}}$ & Cosine denominator and blended-distance normalization & $10^{-8}$ each \\\addlinespace[1.5pt]
$\epsilon_w,\epsilon_g$ & Robot-to-group weights and group-gradient denominators & $10^{-8}$ each \\\addlinespace[1.5pt]
$\epsilon_{\mathrm{pc}}$ & PCGrad denominator; normalized-PCGrad norm guard & $10^{-8}$ each \\\addlinespace[1.5pt]
Encoder stability & Attention normalizers and tanh-bound clipping offset & $10^{-8}$ \\\addlinespace[1.5pt]
$\varepsilon_{\mathrm{min}}$ & Minimum geometry-weighted graph edge length & $10^{-6}$ \\\addlinespace[1.5pt]
Entropy stability & Column denominator and logarithm guards & $10^{-8}$ each \\\addlinespace[1.5pt]
\bottomrule
\end{tabular}
\end{table}

\clearpage
\paperref{Table}{tab:method_settings} identifies which settings in \paperref{Table}{tab:hyperparameters} apply to each method family. Grouping, projection, and architecture rows can be combined to describe the configurations in \paperref{Table}{tab:appendix_full_results}; the two DEG ablations are included separately.

\begin{table}[H]
\centering
\footnotesize\singlespacing
\renewcommand{\arraystretch}{1.15}
\begin{tabular}{@{}>{\raggedright\arraybackslash}p{0.27\textwidth}@{\hspace{0.04\textwidth}}>{\raggedright\arraybackslash}p{0.69\textwidth}@{}}
\toprule
\textbf{Method family} & \textbf{Applicable settings and exceptions} \\
\midrule
Uniform controls & Shared IQL and optimization settings; one actor step per iteration. Group-Uniform samples one fixed group; neither control uses projection or dynamic refreshes. \\\addlinespace[5pt]
Static EG & Default backbone and fixed morphology groups; $M$ actor steps. Refresh and soft-routing settings are inactive. \\\addlinespace[5pt]
Dynamic FGW & FGW coefficient $\alpha$, blend $\alpha_g$, feature scales $b_1,b_2$, and refresh settings $K,W_0,\delta_g$; soft versions additionally use $\kappa$. \\\addlinespace[5pt]
Dynamic Direct & Refresh settings, EMA decay $\lambda$, and medoid-iteration limit; soft versions additionally use $\kappa$. FGW regrouping settings are inactive. \\\addlinespace[5pt]
PCGrad variants & Apply the plain or normalized rule and its numerical guards; Inter-PCGrad additionally projects group aggregates (\ref{app:a_variants}). \\\addlinespace[5pt]
GeoEG variants & Use the selected static or dynamic family's settings. The geometry-weighted matrix defines fixed groups or initializes dynamic groups, as labelled in \paperref{Table}{tab:appendix_full_results}. \\\addlinespace[5pt]
Per-robot heads & Default backbone with robot-specific actor, critic, or both output projections. Shared training settings are retained. \\\addlinespace[5pt]
MoE actor & Expert count and width, gate initialization, and kernel-refresh interval; critics retain the default architecture. \\\addlinespace[5pt]
DEG-Freeze & DEG settings; refreshes stop after the first successful regrouping. \\\addlinespace[5pt]
DEG-Collapse & DEG settings; refreshes continue, but every actor step uses the mean of the group aggregates. \\
\bottomrule
\end{tabular}
\caption{Settings applicable to the evaluated method families. Detailed definitions are given in~\ref{app:a_variants} and~\ref{app:a_ablations}.}
\label{tab:method_settings}
\end{table}

\clearpage
\section{Computational Cost}
\label{app:cost}

\subsection{Actor Parameters and Inference}\label{app:i_parameters}
For the networks in~\ref{app:training}, \paperref{Table}{tab:cost_parameters} counts actor encoders and output projections, excluding value/Q networks, targets, and optimizer state. Shared and MoE counts come from checkpoint shapes. Replacing two shared scalar projections, each with 128 weights and one bias, by 16 robot-specific pairs adds $(16-1)\times2\times(128+1)=3{,}870$ actor parameters. Controls, grouping/projection variants, and ablations retain the shared actor unless an architecture variant is selected. Critic-only heads leave the actor unchanged; full heads have the same actor count as actor-only heads.

\begin{table}[H]
\centering
\footnotesize
\singlespacing
\renewcommand{\arraystretch}{1.15}
\setlength{\tabcolsep}{3pt}
\begin{tabular}{@{}>{\raggedright\arraybackslash}p{0.23\textwidth}@{\hspace{0.04\textwidth}}>{\raggedright\arraybackslash}p{0.31\textwidth}@{\hspace{0.04\textwidth}}>{\raggedright\arraybackslash}p{0.38\textwidth}@{}}
\toprule
\textbf{Configuration} & \textbf{Actor parameters} & \textbf{Inference path} \\
\midrule
Shared backbone & 431,648 & One actor forward pass. \\\addlinespace[5pt]
DEG & 431,648 ($+0$) & Same shared actor pass; group weights are not queried at test time. \\\addlinespace[5pt]
Actor or full per-robot heads & 435,518 ($+3{,}870$; $+0.9\%$) & Select that robot's mean and log-standard-deviation projections. \\\addlinespace[5pt]
MoE & 769,888 ($+78.4\%$) & Evaluate all four experts and combine their outputs with dense gating. \\
\bottomrule
\end{tabular}
\caption{Actor parameter counts and inference computation. Increases are relative to the shared backbone.}
\label{tab:cost_parameters}
\end{table}

\clearpage
\subsection{Training Operations}\label{app:i_runtime}
\paperref{Table}{tab:cost_overhead} lists operations by method family.

\begin{table}[H]
\centering
\footnotesize\singlespacing
\renewcommand{\arraystretch}{1.15}
\begin{tabular}{@{}>{\raggedright\arraybackslash}p{0.27\textwidth}@{\hspace{0.04\textwidth}}>{\raggedright\arraybackslash}p{0.69\textwidth}@{}}
\toprule
\textbf{Method or operation} & \textbf{Training computation} \\
\midrule
Hard grouped updates & One gradient per robot; $M$ unweighted group means and $M$ sequential actor steps. \\\addlinespace[3pt]
Soft grouped updates & One gradient per robot; $M$ weighted aggregates and $M$ sequential actor steps. \\\addlinespace[3pt]
Uniform controls & All robot gradients; average all for Uniform or the selected group for Group-Uniform. One actor step. \\\addlinespace[3pt]
Gradient monitoring & One additional diagnostic pass over all $N=16$ robot gradients every $K$ iterations, including after DEG-Freeze. \\\addlinespace[3pt]
Dynamic FGW refresh & Additional $N$ gradients; if the guard passes, augment cached features, compute pairwise FGW/blended distances, and form Ward groups. Soft versions refresh weights. \\\addlinespace[3pt]
Dynamic Direct refresh & Additional $N$ gradients; if the guard passes, update EMA similarities and balanced $k$-medoids groups. Soft versions refresh weights. \\\addlinespace[3pt]
PCGrad variants & Project within hard groups or across all robots before soft aggregation; Inter-PCGrad also projects group aggregates before stepping. \\\addlinespace[3pt]
GeoEG variants & Fixed or initial groups from the precomputed geometry-weighted matrix; other operations follow the selected base method. \\\addlinespace[3pt]
Per-robot heads & Robot-specific actor and/or critic output projections. \\\addlinespace[3pt]
MoE actor & All four experts per actor pass; periodic QR re-orthogonalization of dense-layer kernels. \\\addlinespace[3pt]
DEG-Freeze & Keep monitoring and $M$ sequential actor steps; stop regrouping after the first successful refresh. \\\addlinespace[3pt]
DEG-Collapse & Keep refreshes; average the $M$ group aggregates and apply that same gradient in $M$ actor steps. \\
\bottomrule
\end{tabular}
\caption{Training operations by method family. Value and twin-Q updates remain global. Mechanism-specific rows combine with the selected hard, soft, or Uniform update rule.}
\label{tab:cost_overhead}
\end{table}

\clearpage
\section*{Declaration of generative AI and AI-assisted technologies in the manuscript preparation process}
During the preparation of this work, the authors used OpenAI Codex (GPT-6) to review the manuscript against the implementation and saved experimental logs and to improve clarity and wording, and to prepare reproducible plotting scripts that derive figures from saved logs. After using this service, the authors reviewed and edited the content as needed and take full responsibility for the content of the published article.

\printbibliography[title={References}]

@inproceedings{abe2024eg,
  author    = {Abe, H. and Osa, T. and Mukuta, Y. and Harada, T.},
  title     = {Cross-Embodiment Offline Reinforcement Learning for Heterogeneous Robot Datasets},
  booktitle = {International Conference on Learning Representations (ICLR)},
  year      = {2026},
  url = {https://proceedings.iclr.cc/paper_files/paper/2026/hash/24a0f56452202e8433d582dbf5095c32-Abstract-Conference.html},
}

@inproceedings{bohlinger2024urma,
  author    = {Bohlinger, N. and Czechmanowski, G. and Krupka, M.~P. and Kicki, P. and Walas, K. and Peters, J. and Tateo, D.},
  title     = {One Policy to Run Them All: an End-to-End Learning Approach to Multi-Embodiment Locomotion},
  booktitle = {Proceedings of The 8th Conference on Robot Learning},
  year = {2025},
  volume = {270},
  pages = {3356--3378},
  url = {https://proceedings.mlr.press/v270/bohlinger25a.html},
  editor = {Agrawal, Pulkit and Kroemer, Oliver and Burgard, Wolfram},
  series = {Proceedings of Machine Learning Research},
  publisher = {PMLR},
}

@inproceedings{brohan2023rt2,
  author = {Brohan, Anthony and Brown, Noah and Carbajal, Justice and Chebotar, Yevgen and Chen, Xi and Choromanski, Krzysztof and Ding, Tianli and Driess, Danny and Dubey, Avinava and Finn, Chelsea and Florence, Pete and Fu, Chuyuan and Gonzalez Arenas, Montse and Gopalakrishnan, Keerthana and Han, Kehang and Hausman, Karol and Herzog, Alexander and Hsu, Jasmine and Ichter, Brian and Irpan, Alex and Joshi, Nikhil and Julian, Ryan and Kalashnikov, Dmitry and Kuang, Yuheng and Leal, Isabel and Lee, Lisa and Lee, Tsang-Wei Edward and Levine, Sergey and Lu, Yao and Michalewski, Henryk and Mordatch, Igor and Pertsch, Karl and Rao, Kanishka and Reymann, Krista and Ryoo, Michael and Salazar, Grecia and Sanketi, Pannag and Sermanet, Pierre and Singh, Jaspiar and Singh, Anikait and Soricut, Radu and Tran, Huong and Vanhoucke, Vincent and Vuong, Quan and Wahid, Ayzaan and Welker, Stefan and Wohlhart, Paul and Wu, Jialin and Xia, Fei and Xiao, Ted and Xu, Peng and Xu, Sichun and Yu, Tianhe and Zitkovich, Brianna},
  title     = {{RT-2}: Vision-Language-Action Models Transfer Web Knowledge to Robotic Control},
  booktitle = {Proceedings of The 7th Conference on Robot Learning},
  year      = {2023},
  volume = {229},
  pages = {2165--2183},
  url = {https://proceedings.mlr.press/v229/zitkovich23a.html},
  editor = {Tan, Jie and Toussaint, Marc and Darvish, Kourosh},
  series = {Proceedings of Machine Learning Research},
  publisher = {PMLR},
}

@inproceedings{doshi2025scaling,
  author    = {Doshi, R. and Walke, H.~R. and Mees, O. and Dasari, S. and Levine, S.},
  title     = {Scaling Cross-Embodied Learning: One Policy for Manipulation, Navigation, Locomotion and Aviation},
  booktitle = {Proceedings of The 8th Conference on Robot Learning},
  year      = {2025},
  volume = {270},
  pages = {496--512},
  url = {https://proceedings.mlr.press/v270/doshi25a.html},
  editor = {Agrawal, Pulkit and Kroemer, Oliver and Burgard, Wolfram},
  series = {Proceedings of Machine Learning Research},
  publisher = {PMLR},
}

@inproceedings{fifty2021tag,
  author    = {Fifty, C. and Amid, E. and Zhao, Z. and Yu, T. and Anil, R. and Finn, C.},
  title     = {Efficiently Identifying Task Groupings for Multi-Task Learning},
  booktitle = {Advances in Neural Information Processing Systems},
  year      = {2021},
  editor = {Ranzato, M. and Beygelzimer, A. and Dauphin, Y. and Liang, P. S. and Wortman Vaughan, J.},
  pages = {27503--27516},
  publisher = {Curran Associates, Inc.},
  url = {https://proceedings.neurips.cc/paper_files/paper/2021/file/e77910ebb93b511588557806310f78f1-Paper.pdf},
  volume = {34},
}

@inproceedings{hendawy2024moore,
  author    = {Hendawy, A. and Peters, J. and D'Eramo, C.},
  title     = {Multi-Task Reinforcement Learning with Mixture of Orthogonal Experts},
  booktitle = {International Conference on Learning Representations (ICLR)},
  year      = {2024},
  url = {https://proceedings.iclr.cc/paper_files/paper/2024/hash/19c145aaad40927c51f4d10eaa339c20-Abstract-Conference.html},
}

@inproceedings{jeong2025stgu,
  author    = {Jeong, W. and Yoon, K.-J.},
  title     = {Selective Task Group Updates for Multi-Task Optimization},
  booktitle = {International Conference on Learning Representations (ICLR)},
  year      = {2025},
  url = {https://proceedings.iclr.cc/paper_files/paper/2025/hash/ed3cd2520148b577039adfade82a5566-Abstract-Conference.html},
}

@inproceedings{kingma2015adam,
  author    = {Kingma, D.~P. and Ba, J.},
  title     = {Adam: A Method for Stochastic Optimization},
  booktitle = {International Conference on Learning Representations (ICLR)},
  year      = {2015},
  url = {https://arxiv.org/abs/1412.6980},
}

@inproceedings{kostrikov2022iql,
  author    = {Kostrikov, I. and Nair, A. and Levine, S.},
  title     = {Offline Reinforcement Learning with Implicit {Q}-Learning},
  booktitle = {International Conference on Learning Representations (ICLR)},
  year      = {2022},
  url = {https://arxiv.org/abs/2110.06169},
}

@inproceedings{li2024gradtag,
  author    = {Li, D. and Sharma, A. and Zhang, H.~R.},
  title     = {Scalable Multitask Learning Using Gradient-Based Estimation of Task Affinity},
  booktitle = {Proceedings of the 30th ACM SIGKDD Conference on Knowledge Discovery and Data Mining (KDD)},
  year      = {2024},
  url = {https://arxiv.org/abs/2409.06091},
  doi = {10.1145/3637528.3671835},
  pages = {1542--1553},
  publisher = {Association for Computing Machinery},
}

@article{lin2026dbmtl,
  author  = {Lin, B. and Jiang, W. and Ye, F. and Zhang, Y. and Chen, P. and Chen, Y.-C. and Liu, S. and Tsang, I.~W. and Kwok, J.~T.},
  title   = {Dual-Balancing for Multi-Task Learning},
  journal = {Neural Networks},
  volume  = {195},
  year    = {2026},
  doi     = {10.1016/j.neunet.2025.108317},
  eid = {108317},
}

@inproceedings{liu2021cagrad,
  author    = {Liu, B. and Liu, X. and Jin, X. and Stone, P. and Liu, Q.},
  title     = {Conflict-Averse Gradient Descent for Multi-Task Learning},
  booktitle = {Advances in Neural Information Processing Systems},
  year      = {2021},
  editor = {Ranzato, M. and Beygelzimer, A. and Dauphin, Y. and Liang, P. S. and Wortman Vaughan, J.},
  pages = {18878--18890},
  publisher = {Curran Associates, Inc.},
  url = {https://proceedings.neurips.cc/paper_files/paper/2021/file/9d27fdf2477ffbff837d73ef7ae23db9-Paper.pdf},
  volume = {34},
}

@inproceedings{octo2024,
  author = {{Octo Model Team} and Ghosh, D. and Walke, H. and Pertsch, K. and Black, K. and Mees, O. and Dasari, S. and Hejna, J. and Kreiman, T. and Xu, C. and Luo, J. and Tan, Y.~L. and Chen, L.~Y. and Sanketi, P. and Vuong, Q. and Xiao, T. and Sadigh, D. and Finn, C. and Levine, S.},
  title   = {Octo: An Open-Source Generalist Robot Policy},
  year    = {2024},
  booktitle = {Proceedings of Robotics: Science and Systems},
  doi = {10.15607/RSS.2024.XX.090},
  url = {https://www.roboticsproceedings.org/rss20/p090.html},
}

@inproceedings{open_x_embodiment2024,
  author = {{Open X-Embodiment Collaboration} and O’Neill, Abby and Rehman, Abdul and Maddukuri, Abhiram and Gupta, Abhishek and Padalkar, Abhishek and Lee, Abraham and Pooley, Acorn and Gupta, Agrim and Mandlekar, Ajay and Jain, Ajinkya and Tung, Albert and Bewley, Alex and Herzog, Alex and Irpan, Alex and Khazatsky, Alexander and Rai, Anant and Gupta, Anchit and Wang, Andrew and Singh, Anikait and Garg, Animesh and Kembhavi, Aniruddha and Xie, Annie and Brohan, Anthony and Raffin, Antonin and Sharma, Archit and Yavary, Arefeh and Jain, Arhan and Balakrishna, Ashwin and Wahid, Ayzaan and Burgess-Limerick, Ben and Kim, Beomjoon and Schölkopf, Bernhard and Wulfe, Blake and Ichter, Brian and Lu, Cewu and Xu, Charles and Le, Charlotte and Finn, Chelsea and Wang, Chen and Xu, Chenfeng and Chi, Cheng and Huang, Chenguang and Chan, Christine and Agia, Christopher and Pan, Chuer and Fu, Chuyuan and Devin, Coline and Xu, Danfei and Morton, Daniel and Driess, Danny and Chen, Daphne and Pathak, Deepak and Shah, Dhruv and Büchler, Dieter and Jayaraman, Dinesh and Kalashnikov, Dmitry and Sadigh, Dorsa and Johns, Edward and Foster, Ethan and Liu, Fangchen and Ceola, Federico and Xia, Fei and Zhao, Feiyu and Stulp, Freek and Zhou, Gaoyue and Sukhatme, Gaurav S. and Salhotra, Gautam and Yan, Ge and Feng, Gilbert and Schiavi, Giulio and Berseth, Glen and Kahn, Gregory and Wang, Guanzhi and Su, Hao and Fang, Hao-Shu and Shi, Haochen and Bao, Henghui and Ben Amor, Heni and Christensen, Henrik I and Furuta, Hiroki and Walke, Homer and Fang, Hongjie and Ha, Huy and Mordatch, Igor and Radosavovic, Ilija and Leal, Isabel and Liang, Jacky and Abou-Chakra, Jad and Kim, Jaehyung and Drake, Jaimyn and Peters, Jan and Schneider, Jan and Hsu, Jasmine and Bohg, Jeannette and Bingham, Jeffrey and Wu, Jeffrey and Gao, Jensen and Hu, Jiaheng and Wu, Jiajun and Wu, Jialin and Sun, Jiankai and Luo, Jianlan and Gu, Jiayuan and Tan, Jie and Oh, Jihoon and Wu, Jimmy and Lu, Jingpei and Yang, Jingyun and Malik, Jitendra and Silvério, João and Hejna, Joey and Booher, Jonathan and Tompson, Jonathan and Yang, Jonathan and Salvador, Jordi and Lim, Joseph J. and Han, Junhyek and Wang, Kaiyuan and Rao, Kanishka and Pertsch, Karl and Hausman, Karol and Go, Keegan and Gopalakrishnan, Keerthana and Goldberg, Ken and Byrne, Kendra and Oslund, Kenneth and Kawaharazuka, Kento and Black, Kevin and Lin, Kevin and Zhang, Kevin and Ehsani, Kiana and Lekkala, Kiran and Ellis, Kirsty and Rana, Krishan and Srinivasan, Krishnan and Fang, Kuan and Singh, Kunal Pratap and Zeng, Kuo-Hao and Hatch, Kyle and Hsu, Kyle and Itti, Laurent and Chen, Lawrence Yunliang and Pinto, Lerrel and Fei-Fei, Li and Tan, Liam and Fan, Linxi Jim and Ott, Lionel and Lee, Lisa and Weihs, Luca and Chen, Magnum and Lepert, Marion and Memmel, Marius and Tomizuka, Masayoshi and Itkina, Masha and Castro, Mateo Guaman and Spero, Max and Du, Maximilian and Ahn, Michael and Yip, Michael C. and Zhang, Mingtong and Ding, Mingyu and Heo, Minho and Srirama, Mohan Kumar and Sharma, Mohit and Kim, Moo Jin and Kanazawa, Naoaki and Hansen, Nicklas and Heess, Nicolas and Joshi, Nikhil J and Suenderhauf, Niko and Liu, Ning and Di Palo, Norman and Shafiullah, Nur Muhammad Mahi and Mees, Oier and Kroemer, Oliver and Bastani, Osbert and Sanketi, Pannag R and Miller, Patrick Tree and Yin, Patrick and Wohlhart, Paul and Xu, Peng and Fagan, Peter David and Mitrano, Peter and Sermanet, Pierre and Abbeel, Pieter and Sundaresan, Priya and Chen, Qiuyu and Vuong, Quan and Rafailov, Rafael and Tian, Ran and Doshi, Ria and Martín-Martín, Roberto and Baijal, Rohan and Scalise, Rosario and Hendrix, Rose and Lin, Roy and Qian, Runjia and Zhang, Ruohan and Mendonca, Russell and Shah, Rutav and Hoque, Ryan and Julian, Ryan and Bustamante, Samuel and Kirmani, Sean and Levine, Sergey and Lin, Shan and Moore, Sherry and Bahl, Shikhar and Dass, Shivin and Sonawani, Shubham and Song, Shuran and Xu, Sichun and Haldar, Siddhant and Karamcheti, Siddharth and Adebola, Simeon and Guist, Simon and Nasiriany, Soroush and Schaal, Stefan and Welker, Stefan and Tian, Stephen and Ramamoorthy, Subramanian and Dasari, Sudeep and Belkhale, Suneel and Park, Sungjae and Nair, Suraj and Mirchandani, Suvir and Osa, Takayuki and Gupta, Tanmay and Harada, Tatsuya and Matsushima, Tatsuya and Xiao, Ted and Kollar, Thomas and Yu, Tianhe and Ding, Tianli and Davchev, Todor and Zhao, Tony Z. and Armstrong, Travis and Darrell, Trevor and Chung, Trinity and Jain, Vidhi and Vanhoucke, Vincent and Zhan, Wei and Zhou, Wenxuan and Burgard, Wolfram and Chen, Xi and Wang, Xiaolong and Zhu, Xinghao and Geng, Xinyang and Liu, Xiyuan and Liangwei, Xu and Li, Xuanlin and Lu, Yao and Ma, Yecheng Jason and Kim, Yejin and Chebotar, Yevgen and Zhou, Yifan and Zhu, Yifeng and Wu, Yilin and Xu, Ying and Wang, Yixuan and Bisk, Yonatan and Cho, Yoonyoung and Lee, Youngwoon and Cui, Yuchen and Cao, Yue and Wu, Yueh-Hua and Tang, Yujin and Zhu, Yuke and Zhang, Yunchu and Jiang, Yunfan and Li, Yunshuang and Li, Yunzhu and Iwasawa, Yusuke and Matsuo, Yutaka and Ma, Zehan and Xu, Zhuo and Cui, Zichen Jeff and Zhang, Zichen and Lin, Zipeng},
  title = {{Open X-Embodiment}: Robotic Learning Datasets and {RT-X} Models},
  year    = {2024},
  booktitle = {2024 IEEE International Conference on Robotics and Automation (ICRA)},
  pages = {6892--6903},
  doi = {10.1109/ICRA57147.2024.10611477},
}

@online{pascal2021improved,
  author  = {Pascal, L. and Michiardi, P. and Bost, X. and Huet, B. and Zuluaga, M.~A.},
  title   = {Improved Optimization Strategies for Deep Multi-Task Networks},
  year    = {2021},
  doi = {10.48550/arXiv.2109.11678},
  titleaddon = {Preprint},
  organization = {arXiv},
}

@inproceedings{springenberg2024offline,
  author    = {Springenberg, J.~T. and Abdolmaleki, A. and Zhang, J. and Groth, O. and Bloesch, M. and Lampe, T. and Brakel, P. and Bechtle, S.~M.~E. and Kapturowski, S. and Hafner, R. and Heess, N. and Riedmiller, M.},
  title     = {Offline Actor-Critic Reinforcement Learning Scales to Large Models},
  booktitle = {Proceedings of the 41st International Conference on Machine Learning},
  year      = {2024},
  volume = {235},
  pages = {46323--46350},
  url = {https://proceedings.mlr.press/v235/springenberg24a.html},
  editor = {Salakhutdinov, Ruslan and Kolter, Zico and Heller, Katherine and Weller, Adrian and Oliver, Nuria and Scarlett, Jonathan and Berkenkamp, Felix},
  series = {Proceedings of Machine Learning Research},
  publisher = {PMLR},
}

@inproceedings{todorov2012mujoco,
  author    = {Todorov, E. and Erez, T. and Tassa, Y.},
  title     = {{MuJoCo}: A Physics Engine for Model-Based Control},
  booktitle = {IEEE/RSJ International Conference on Intelligent Robots and Systems (IROS)},
  year      = {2012},
  pages = {5026--5033},
  doi = {10.1109/IROS.2012.6386109},
}

@inproceedings{vayer2019fgw,
  author    = {Vayer, T. and Chapel, L. and Flamary, R. and Tavenard, R. and Courty, N.},
  title     = {Optimal Transport for Structured Data with Application on Graphs},
  booktitle = {Proceedings of the 36th International Conference on Machine Learning},
  year      = {2019},
  volume = {97},
  pages = {6275--6284},
  url = {https://proceedings.mlr.press/v97/titouan19a.html},
  editor = {Chaudhuri, Kamalika and Salakhutdinov, Ruslan},
  series = {Proceedings of Machine Learning Research},
  publisher = {PMLR},
}

@inproceedings{yang2025multiloco,
  author  = {Yang, S. and Fu, Z. and Cao, Z. and Guo, J. and Wensing, P. and Zhang, W. and Chen, H.},
  title = {{Multi-Loco}: Unifying Multi-Embodiment Legged Locomotion via Reinforcement Learning Augmented Diffusion},
  year    = {2025},
  booktitle = {Proceedings of The 9th Conference on Robot Learning},
  volume = {305},
  pages = {1030--1048},
  url = {https://proceedings.mlr.press/v305/yang25a.html},
  editor = {Lim, Joseph and Song, Shuran and Park, Hae-Won},
  series = {Proceedings of Machine Learning Research},
  publisher = {PMLR},
}

@inproceedings{yu2020pcgrad,
  author    = {Yu, T. and Kumar, S. and Gupta, A. and Levine, S. and Hausman, K. and Finn, C.},
  title     = {Gradient Surgery for Multi-Task Learning},
  booktitle = {Advances in Neural Information Processing Systems},
  year      = {2020},
  editor = {Larochelle, H. and Ranzato, M. and Hadsell, R. and Balcan, M. F. and Lin, H.},
  pages = {5824--5836},
  publisher = {Curran Associates, Inc.},
  url = {https://proceedings.neurips.cc/paper_files/paper/2020/file/3fe78a8acf5fda99de95303940a2420c-Paper.pdf},
  volume = {33},
}

@article{dijkstra1959note,
  author  = {Dijkstra, E. W.},
  title   = {A Note on Two Problems in Connexion with Graphs},
  journal = {Numerische Mathematik},
  volume  = {1},
  pages   = {269--271},
  year    = {1959},
  doi     = {10.1007/BF01386390},
  number = {1},
}

@book{kaufman1990finding,
  author    = {Kaufman, L. and Rousseeuw, P. J.},
  title     = {Finding Groups in Data: An Introduction to Cluster Analysis},
  publisher = {John Wiley \& Sons},
  address   = {New York},
  year      = {1990},
  doi       = {10.1002/9780470316801}
}

@article{ward1963hierarchical,
  author  = {Ward, Jr., J. H.},
  title   = {Hierarchical Grouping to Optimize an Objective Function},
  journal = {Journal of the American Statistical Association},
  volume  = {58},
  number  = {301},
  pages   = {236--244},
  year    = {1963},
  doi     = {10.1080/01621459.1963.10500845}
}

@article{shannon1948i,
  author  = {Shannon, C. E.},
  title   = {A Mathematical Theory of Communication},
  journal = {The Bell System Technical Journal},
  volume  = {27},
  number  = {3},
  pages   = {379--423},
  year    = {1948},
  doi     = {10.1002/j.1538-7305.1948.tb01338.x}
}
\end{document}